\pdfoutput=1
\documentclass[11pt]{article}

\usepackage[final]{acl}

\usepackage{times}
\usepackage{latexsym}
\usepackage[T1]{fontenc}
\usepackage[utf8]{inputenc}
\usepackage{microtype}
\usepackage{inconsolata}

\usepackage{graphicx}
\usepackage{caption}
\usepackage{float}

\usepackage{amsmath}
\usepackage{multirow}
\usepackage{booktabs}
\usepackage{adjustbox} 
\usepackage{enumitem}
\usepackage{gensymb}

\usepackage{wrapfig}

\usepackage{algorithm}
\usepackage{algpseudocode}

\usepackage{siunitx}
\usepackage[capitalize]{cleveref}

\usepackage{xcolor}
\usepackage[most]{tcolorbox}
\tcbuselibrary{listingsutf8, skins}

\newtcolorbox{UserPrompt}{
  enhanced,
  colback=gray!10,
  colframe=blue!80!black,
  coltitle=white,
  fonttitle=\bfseries,
  title=User Prompt,
  arc=3pt,
  boxrule=1pt,
  left=2mm,
  right=2mm,
  top=1mm,
  bottom=1mm
}

\newtcolorbox{SystemPrompt}{
  enhanced,
  colback=green!5!white,
  colframe=green!75!black,
  coltitle=white,
  fonttitle=\bfseries,
  title=System Prompt,
  arc=3pt,
  boxrule=1pt,
  left=2mm,
  right=2mm,
  top=1mm,
  bottom=1mm
}

\setlist[itemize]{
  itemsep=1pt,     
  parsep=0pt,     
  topsep=0pt,      
  partopsep=0pt,   
  leftmargin=*     
}

\newcommand{\myparagraph}[1]{\vspace{0.25em}\noindent \textbf{#1}\hspace{0.5em}}

\newcommand{\rotlarge}{\textsc{RotBench-large}}
\newcommand{\rotsmall}{\textsc{RotBench-small}}
\newcommand{\rotbench}{\textsc{RotBench}}

\newcommand{\gptfour}{GPT-4o}
\newcommand{\gptfourone}{GPT-4.1}
\newcommand{\gptfive}{GPT-5}
\newcommand{\othree}{o3}
\newcommand{\geminitwoflash}{Gemini-2.0-Flash}
\newcommand{\geminitwofiveflash}{Gemini-2.5-Flash}
\newcommand{\geminitwofivepro}{Gemini-2.5-Pro}
\newcommand{\llama}{Llama-3.2-11B-Instruct}
\newcommand{\qwen}{Qwen-2.5-VL-7B-Instruct}
\newcommand{\claude}{Claude-4.1-Opus}
\newcommand{\gglphi}{Phi-4-MM-Instruct}
\newcommand{\gemma}{Gemma-3-12B-Instruct}
\newcommand{\smallqwen}{Qwen-2.5-VL-3B-Instruct}
\newcommand{\molmo}{Molmo-7B-O}

\title{RotBench: Evaluating Multimodal Large Language Models on \\ Identifying Image Rotation}

\author{
  \textbf{Tianyi Niu$^1$ \quad Jaemin Cho$^2$ \quad Elias Stengel-Eskin$^3$ \quad Mohit Bansal}$^1$  \\ [1ex]
  $^1$UNC Chapel Hill$\quad$$^2$Allen Institute for Artificial Intelligence \\$^3$The University of Texas at Austin
}

\begin{document}
\maketitle

\begin{abstract}
 
We investigate to what extent Multimodal Large Language Models (MLLMs) can accurately identify the orientation of input images rotated 0°, 90°, 180°, and 270°. This task demands robust visual reasoning capabilities to detect rotational cues and contextualize spatial relationships within images, regardless of their orientation. To evaluate MLLMs on these abilities, we introduce \rotbench{}, a 350-image manually-filtered benchmark comprising lifestyle, portrait, and landscape images. Despite the relatively simple nature of this task, we show that several state-of-the-art open and proprietary MLLMs, including \gptfive{}, \othree{}, and \geminitwofivepro{}, do not reliably identify rotation in input images. Providing models with auxiliary information---including captions, depth maps, and more---or using chain-of-thought prompting offers only small and inconsistent improvements. Our results indicate that most models are able to reliably identify right-side-up (0°) images, while certain models are able to identify upside-down (180°) images. None can reliably distinguish between 90° and 270° rotated images. Simultaneously showing the image rotated in different orientations leads to moderate performance gains for reasoning models, while a modified setup using voting improves the performance of weaker models. We further show that fine-tuning does not improve models' ability to distinguish 90° and 270° rotations, despite substantially improving the identification of 180° images. Together, these results reveal a significant gap between MLLMs' spatial reasoning capabilities and human perception in identifying rotation.\footnote{Code and data: \href{https://github.com/tianyiniu/RotBench}{https://github.com/tianyiniu/RotBench}}

\end{abstract}

\section{Introduction}
\label{sec:intro}

\begin{figure}[ht]
    \centering
    \includegraphics[width=0.45\textwidth]{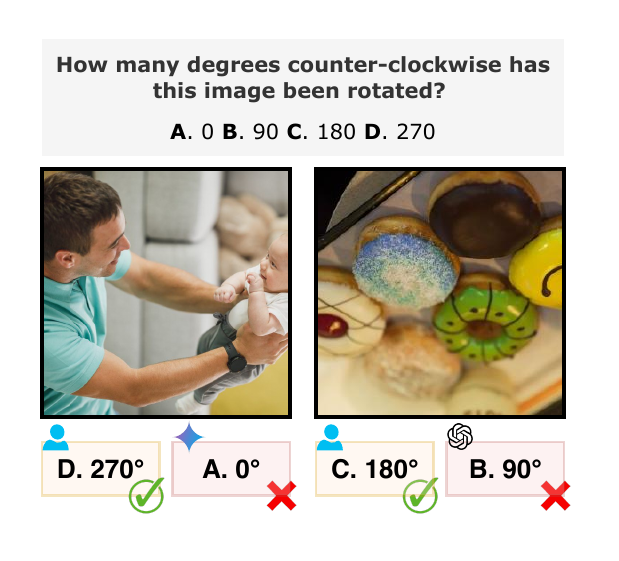}
    \vspace{-0.5em}
    \caption{We present two \rotbench{} images: one (left) to \geminitwofivepro{}, the other (right) to \gptfive{}. Humans can easily identify the correct rotation of the two images, but both models fail to do so.}
    \vspace{-1em}
    \label{fig:img_mini}
\end{figure}

\begin{figure*}[htbp]
    \centering
    \includegraphics[width=\textwidth]{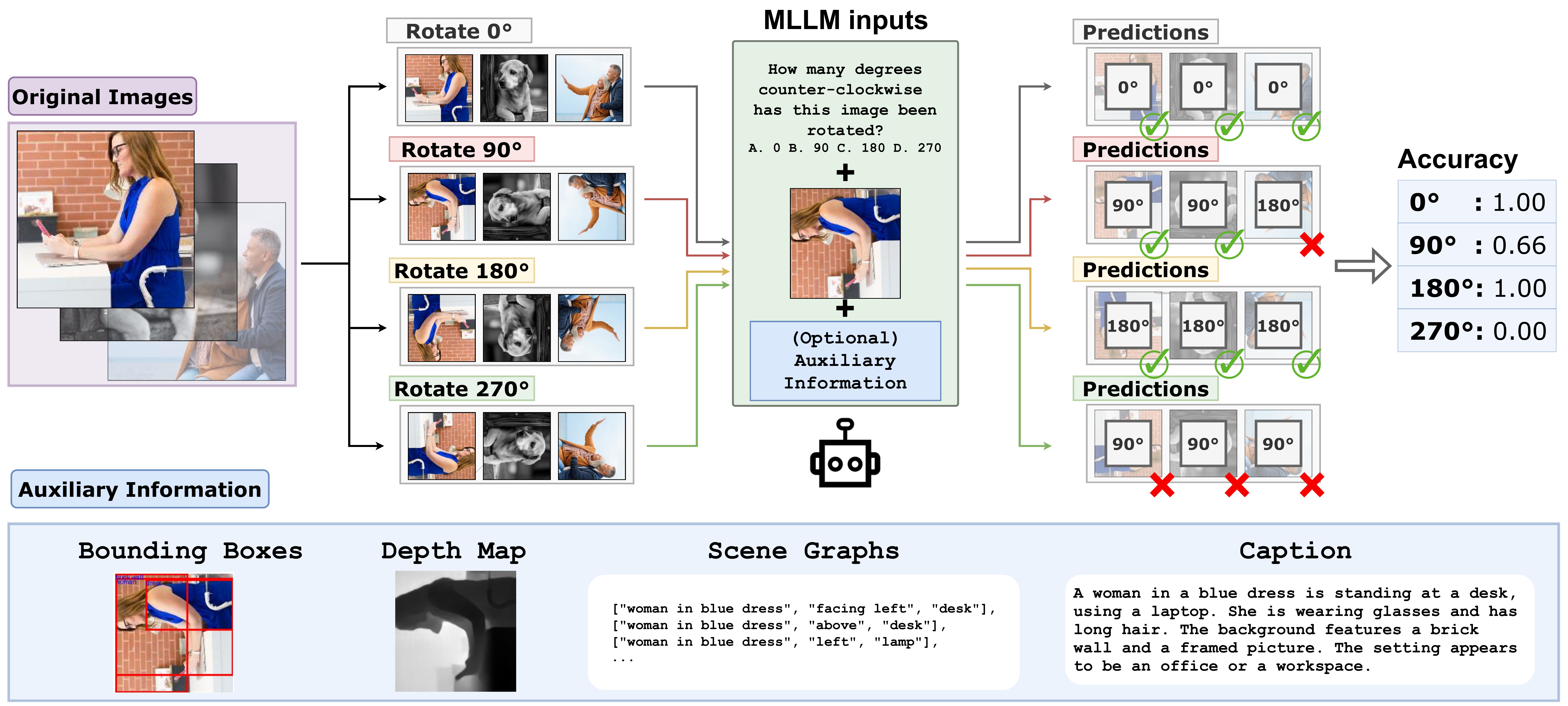}
    \caption{\textbf{\rotbench{}} evaluation pipeline: for each image in \rotbench{}, we rotate the image 0°, 90°, 180°, and 270° counter-clockwise. We represent the rotation estimation problem as a multiple-choice question answering problem  (\cref{sec:app_prompts}), and separately measure accuracy on each image orientation. We optionally provide different forms of auxiliary information to aid the model in identifying image rotation. We emphasize that all forms of auxiliary information are separately extracted for each rotation; the ground truth rotation is not marked.}
    \label{fig:main_pipeline}
\end{figure*}

Advancements in Multimodal Large Language Models (MLLMs) have led to increased performance in complex visual tasks, such as image-text retrieval, image segmentation, and visual question answering \cite{li2025surveystateartlarge, chen2023, ravi2024sam2, fu2024mme, chen2023pali, Qwen2.5-VL, openai2025o3, gemini2025gemini2.5, liu2023llava, deitke2024molmo}. However, a growing body of recent work suggests that MLLMs are sensitive to simple image transformations \cite{anis2025}, such as rotations, flips, and blurs, and they fail on tasks that are intuitive to humans \cite{fu2024blink, pothiraj2025captureevaluatingspatialreasoning, tong2024eyeswideshutexploring}. Downstream tasks involving a rotating camera, such as robotic arm manipulation or first-person extreme sports analysis, require MLLMs to demonstrate robust spatial reasoning, regardless of image orientation \textcolor{black}{(\cref{sec:downstream_apps})}. Given these challenges, this work explores a fundamental question: can MLLMs identify image orientation?  

Humans can quickly recognize whether an image has been rotated \cite{Shepard1971, Vandenberg1978}; for example, it is easy for us to recognize that the left image in \cref{fig:img_mini} is not upright. A human viewer can use the orientation of the couch in the background to infer that the father is actually lying on his back rather than standing up. The simple task of identifying image rotation requires reconciling the image's subjects, background, and semantics. Here, we show that identifying rotation remains a challenge, even in frontier MLLMs.

We introduce \rotbench{} (\cref{sec:dataset}), a benchmark for evaluating MLLMs' ability to recognize rotation in images. \rotbench{} consists of images sampled from Spatial-MM \cite{shiri2024empirical} and is comprised of two subsets: the 300-image \rotlarge{} and the 50-image \rotsmall{}. \rotbench{} is carefully constructed to be challenging but fair, with a two-stage filtering procedure to remove images that are indistinguishable under different degrees of rotation. \Cref{sec:dataset} describes our dataset construction. 

Using \rotbench{}, we explore whether frontier MLLMs can identify rotation in input images rotated 0°, 90°, 180°, and 270° (\cref{fig:main_pipeline}). We also evaluate whether providing various forms of auxiliary information or using chain-of-thought \cite{wei2022cot} prompting improves performance (\cref{sec:additional_info}). We find that models are able to consistently identify right-side-up (0°) images. However, only stronger models are able to identify upside-down (180°) images. All models fail to accurately distinguish between 90° and 270° images (\cref{sec:main_results}, \cref{sec:confusion_matrix}, \cref{sec:cw_ccw_experiment}). We find adding auxiliary information offers minimal and inconsistent improvement, often improving performance on 270° images at the expense of 90° images. 

Leveraging MLLMs' tendency to accurately identify 0° images, we attempt to improve performance by \textit{further} rotating an input image 0°, 90°, 180°, 270° and presenting MLLMs with all rotations simultaneously (\cref{sec:main_results}). Using this approach, reasoning models show performance improvements, while weaker models see performance degradation. Further extending this idea, we utilize a voting approach to algebraically obtain a majority vote for the correct ground truth orientation. We obtain a model prediction for each further rotation by subtracting the added angle. While this approach does show significant performance improvements on weaker models, it lacks scalability as it requires multiple model calls for each prediction, and assumes \textit{a priori} knowledge of all possible orientations (\cref{sec:majority_vote_exp}).

Finally, fine-tuning on out-of-domain data (\cref{sec:ft_experiment}) significantly improved performance on identifying 180° images but not 90° and 270°. Interestingly, we find an oscillating pattern of performance changes between 90° and 270° as training progresses. Any performance improvement in 90° is matched with a degradation in performance of 270° and vice versa. These results suggest the presence of two local optima hindering across-the-board progress.  

Our findings demonstrate the tested MLLMs significantly underperform compared to humans when it comes to spatial reasoning involving rotation detection, highlighting the need for integrating rotation-awareness into modern training pipelines.

\section{Related Work}
\label{sec:related_work}

\textcolor{black}{This section reviews the literature most closely related to our work. Additional relevant areas are discussed in \cref{sec:further_relevant_work}.}

\myparagraph{Image orientation estimation.}
Fine-tuning models to identify image orientation has been the focus of prior work \cite{xu2024survey}. For example, \citet{Fischer2015image} and \citet{Joshi2017} focused on fine-tuning convolutional neural networks (CNNs) to estimate and identify image rotation. 
While our work tackles a similar task, we are instead interested in the problem as a test of general-purpose MLLMs' inherent reasoning abilities, i.e., whether they can estimate image rotation without extensive fine-tuning. Note that our work aligns more closely with the problem of \textit{image}, not \textit{camera}, orientation estimation. \textcolor{black}{We further discuss camera orientation estimation in \cref{sec:further_relevant_work}.}

\myparagraph{Spatial reasoning in MLLMs.}
Beyond robustness, spatial relation understanding is a notable weakness of current MLLMs. \citet{kamath2023whatsup} curate the What’s Up benchmark to isolate “left/right/above/below” relations, showing a significant gap in performance between humans and MLLMs.
\citet{shiri2024empirical} further develop the Spatial-MM dataset and demonstrate that providing bounding boxes or scene graphs yields only modest gains. Both illustrate that MLLMs struggle with certain challenging cross-modal spatial reasoning tasks.

\myparagraph{Gap between human perception and MLLMs.}
A growing body of work shows MLLMs exhibit fundamental gaps compared to human perceptual capabilities. \citet{pothiraj2025captureevaluatingspatialreasoning} propose CAPTURe, a benchmark for occluded object counting, and report sharp drops in model accuracy on both synthetic and real images. \citet{zhou2025mmvm} proposes MMVM, a benchmark for visual matching across images. \citet{fu2024blink} collect BLINK, a dataset comprised of visual tasks humans can solve in a `blink,' such as identifying visual similarity and relative depth. Both \citet{zhou2025mmvm} and \citet{fu2024blink} report low zero-shot accuracy on their respective tasks, suggesting MLLMs lack many of the intuitive reasoning mechanisms that underpin human visual perception. In this vein, our work provides a novel perspective to analyze and interpret the spatial reasoning capabilities of MLLMs, with results indicating that models struggle with this previously underexplored challenge.

\section{\rotbench{}}
\label{sec:dataset}

We introduce \textbf{\rotbench{}}, a benchmark for evaluating models' ability to identify rotation in input images. \rotbench{} is created using images from Spatial-MM \cite{shiri2024empirical} and includes two subsets: the 300-image \textbf{\rotlarge{}} and the 50-image \textbf{\rotsmall{}}. While rotating an image is straightforward, not all images are meaningful under rotation. A rotated portrait of a human, such as the image shown in \cref{fig:main_pipeline}, will be easily noticed by human viewers. However, a top-down view of a simple tabletop does not significantly differ when rotated (\cref{fig:appendix_sample_imgs}). We use a two-stage filtering process to ensure different rotations of each image are clearly distinguishable. This section provides an overview of our filtering procedure. \Cref{sec:data_details} describes our dataset procedure and statistics in further detail.   

\myparagraph{Stage 1.}
We randomly sample 300 images from Spatial-MM \cite{shiri2024empirical}. Stage 1 involves a single annotator. Depending on the amount of visual signals available, the Stage 1 annotator decides to either accept, discard, or flag each image. Flagged images then proceed to Stage 2. We provide further examples and details of accepted, flagged, and discarded images in \cref{fig:appendix_sample_imgs}.

\myparagraph{Stage 2.}
Stage 2 involves a group of three human evaluators. Each flagged image is rotated 0°, 90°, 180°, and 270° counter-clockwise and then presented to the evaluators as multiple-choice questions (\cref{sec:annotator_interface}). During Stage 2, an evaluator will see each image four times, once per orientation. Any image that elicits an incorrect answer from two or more evaluators across all four orientations is discarded. Otherwise, the image is accepted. Stages 1 and 2 are repeated until 300 images are accepted. 

\myparagraph{\rotlarge{} and \rotsmall{}.}
All images that have been accepted in Stages 1 and 2 are organized into \rotlarge{}. As each image is rotated in four orientations, obtaining human performance on \rotlarge{} is costly. Fortunately, Stage 2 provides a human baseline for the subset of \rotlarge{} images that have been flagged (25 images). We expand this subset by further sampling images from Spatial-MM. From these additional images, we only select images that fit the criteria for flagging to proceed to another round of Stage 2 evaluation. This process repeats until we reach a total of 50 images, organizing them into \rotsmall{}.\footnote{The human evaluators all exhibit high accuracy, averaging > 0.97 for all rotations.}

\section{Experiment Setup}
\label{sec:experiments}

\subsection{Models}
\label{sec:models}

We evaluate various open-weight and proprietary MLLMs on \rotbench{}: \qwen{} \cite{Qwen2.5-VL}; \gptfour{} \cite{openai2024gpt4ocard}, \gptfourone{} \cite{openai2025gpt41}, \othree{} \cite{openai2025o3}, \gptfive{} \cite{openai2025gpt5}, \geminitwoflash{} \cite{google2024gemini20}, \geminitwofiveflash{} \cite{gemini2025gemini2.5}, and \geminitwofivepro{} \cite{gemini2025gemini2.5}. Due to cost and resource limitations, we evaluate \geminitwofiveflash{}, \geminitwofivepro{}, \gptfourone{}, \gptfive{}, and \othree{} on \rotsmall{}. Responses are obtained through greedy decoding, while all chain-of-thought \cite{wei2022cot} responses are obtained with a temperature of 0.3.\footnote{We perform an ablation study where we vary the sampling temperature (\Cref{tab:temp_ablation}). We use the default sampling temperature for proprietary models that do not expose a temperature parameter in their API interface.} \textcolor{black}{In addition, we provide further results on \llama{} \cite{grattafiori2024llama3herdmodels}, \smallqwen{} \cite{Qwen2.5-VL}, \gglphi{} \cite{phi4}, \gemma{} \cite{gemma3}, \molmo{} \cite{molmo}, and \claude{} \cite{claude} in \cref{sec:additional_results}.}

\subsection{Setup and Evaluation}
\label{sec:setup_evaluation}

We rotate each image in \rotlarge{} and \rotsmall{} by 0°, 90°, 180°, and 270° counter-clockwise, resulting in a total of 1,200 and 200 images.
Note that a 90° counter-clockwise rotation is equivalent to a 270° clockwise rotation.
For each image and orientation, we provide the model with the image, a brief description of the task, and various forms of auxiliary information (\cref{sec:additional_info}). We frame this task as a four-way classification problem. To ensure robustness, the mapping between letter choice and degree of rotation is randomized for each prompt. We evaluate models on \rotlarge{} and report average accuracy and standard deviation across 3 runs in \cref{tab:large_evals}. We use the same procedure, albeit with only 2 runs, on \rotsmall{} and report results in \cref{tab:small_evals}. All prompts used are given in \cref{sec:app_prompts}. \textcolor{black}{We also present evaluations under a regression formulation and a multi-choice setup with increased granularity in \cref{sec:alt_setups}, all leading to inferior performance. We select our current setup for its simplicity and because it already challenges frontier models.}

\subsection{Auxiliary Information}
\label{sec:additional_info}

\Cref{fig:appendix_aux_info} illustrates all forms of auxiliary information provided to the model. Note that all auxiliary information is separately extracted for each rotation, ensuring our approach does not depend on prior knowledge of the image's orientation.

\myparagraph{Captions.}
For each image and rotation, we instruct \gptfour{} to provide a detailed caption (\cref{sec:app_prompts}). We emphasize that each image is captioned four times, once per rotation.

\myparagraph{Bounding Boxes.}
For each image and rotation, we first use \gptfour{} to extract the primary subjects within the image (\cref{sec:app_prompts}). Next, along with the image, the list of subjects is given to GroundingDINO \cite{liu2023grounding} to extract a set of normalized coordinates for each subject,\footnote{Each set of coordinates is a four-element tuple, composed of $\texttt{[x\_min, y\_min, x\_max, y\_max]}$.} which is directly injected into the prompt.

\myparagraph{Scene Graphs.}
A scene graph \cite{zhu2022scenegraphgenerationcomprehensive} codifies relationships between objects in an image as a three-element tuple [object 1, predicate, object 2]. Using the extracted subjects from the previous section, we prompt \gptfour{} to generate a scene graph for the image.

\myparagraph{Depth Maps.}
We obtain depth maps for each image using ZoeDepth \cite{bhat2023zoedepth}. Rather than rotating the depth map obtained from 0°, we separately obtain depth maps for all four rotations.

\myparagraph{Segmentation Maps.}
Using the previously extracted bounding boxes, we obtain a segmentation map of each image and orientation using SAM 2 \cite{ravi2024sam2}.

\myparagraph{Chain-of-Thought.}
To evaluate whether our multiple-choice setup is hampering performance on this task, we modify the prompt to encourage the model to produce reasoning chains instead of a single letter choice.

\myparagraph{Rotation Grid.}
We test if explicitly allowing models to ``visualize'' rotations aid performance by providing the input image along with three copies of the image further rotated 90°, 180°, 270°. We compose these four images into a single \textit{rotation grid}. Each image is captioned with the degree of further rotation, independent of the ground truth rotation. We provide a further experiment (\textit{rotation grid guided}) where we explicitly prompt the model to identify an "anchor" image and algebraically calculate the original image's ground truth rotation. All rotation grid experiments use CoT prompting.

\section{Main Results}
\label{sec:main_results}

\begin{table*}[ht]
\centering
\small 
\begin{tabular*}{0.85\textwidth}{@{\extracolsep{\fill}} llcccc}
\toprule
\multicolumn{2}{c}{} & \multicolumn{4}{c}{\textbf{Accuracy on Different Degrees of Rotation}} \\
\cmidrule(lr){3-6}
\multicolumn{2}{c}{\textbf{Model}} & \textbf{0°} & \textbf{90°} & \textbf{180°} & \textbf{270°} \\ 
\midrule
{\textit{Qwen-2.5-VL-7B-Instruct}} \\
Zero-shot & & 
$0.99 {\scriptscriptstyle \pm 0.00}$ & 
$0.51 {\scriptscriptstyle \pm 0.01}$ & 
$0.05 {\scriptscriptstyle \pm 0.01}$ & 
$0.09 {\scriptscriptstyle \pm 0.01}$ \\ 
\hspace{3mm} + Caption & & 
$\textbf{1.00} {\scriptscriptstyle \pm 0.00}$ & 
$0.51 {\scriptscriptstyle \pm 0.01}$ & 
$0.23 {\scriptscriptstyle \pm 0.01}$ & 
$0.07 {\scriptscriptstyle \pm 0.00}$ \\ 
\hspace{3mm} + Bounding Box & & 
$0.90 {\scriptscriptstyle \pm 0.00}$ & 
$0.48 {\scriptscriptstyle \pm 0.01}$ & 
$0.01 {\scriptscriptstyle \pm 0.00}$ &
$0.11 {\scriptscriptstyle \pm 0.00}$ \\ 
\hspace{3mm} + Scene Graph & & 
$0.97 {\scriptscriptstyle \pm 0.01}$ & 
$0.51 {\scriptscriptstyle \pm 0.01}$ & 
$0.01 {\scriptscriptstyle \pm 0.01}$ & 
$0.11 {\scriptscriptstyle \pm 0.02}$ \\ 
\hspace{3mm} + Depth Map & & 
$0.93 {\scriptscriptstyle \pm 0.01}$ & 
$0.55 {\scriptscriptstyle \pm 0.02}$ & 
$0.04 {\scriptscriptstyle \pm 0.01}$ & 
$0.13 {\scriptscriptstyle \pm 0.02}$ \\ 
\hspace{3mm} + Segmentation Map & & 
$0.81 {\scriptscriptstyle \pm 0.01}$ & 
$\textbf{0.63} {\scriptscriptstyle \pm 0.02}$ & 
$0.03 {\scriptscriptstyle \pm 0.02}$ & 
$0.16 {\scriptscriptstyle \pm 0.01}$ \\ 
\hspace{3mm} + Chain-of-Thought  & & 
$0.88 {\scriptscriptstyle \pm 0.01}$ &
$0.26 {\scriptscriptstyle \pm 0.02}$ & 
$\textbf{0.34} {\scriptscriptstyle \pm 0.01}$ & 
$0.23 {\scriptscriptstyle \pm 0.02}$ \\ 
\hspace{3mm} + Rotation Grid & & 
$0.57 {\scriptscriptstyle \pm 0.04}$ & 
$0.15 {\scriptscriptstyle \pm 0.02}$ & 
$0.13 {\scriptscriptstyle \pm 0.01}$ & 
$0.28 {\scriptscriptstyle \pm 0.00}$ \\ 
\hspace{3mm} + Rotation Grid Guided & & 
$0.59 {\scriptscriptstyle \pm 0.01}$ & 
$0.12 {\scriptscriptstyle \pm 0.01}$ & 
$0.13 {\scriptscriptstyle \pm 0.00}$ & 
$0.30 {\scriptscriptstyle \pm 0.02}$ \\ 
\hspace{3mm} + \textit{all above} & & 
$0.47 {\scriptscriptstyle \pm 0.03}$ & 
$0.26 {\scriptscriptstyle \pm 0.01}$ & 
$0.17 {\scriptscriptstyle \pm 0.01}$ & 
$\textbf{0.33} {\scriptscriptstyle \pm 0.02}$ \\ 
\midrule
{\textit{\geminitwofiveflash{}}} \\ 
Zero-shot & & 
$\textbf{1.00} {\scriptscriptstyle \pm 0.00}$ & 
$0.30 {\scriptscriptstyle \pm 0.00}$ & 
$0.72 {\scriptscriptstyle \pm 0.00}$ & 
$0.44 {\scriptscriptstyle \pm 0.01}$ \\ 
\hspace{3mm} + Caption & & 
$\textbf{1.00} {\scriptscriptstyle \pm 0.00}$ & 
$0.38 {\scriptscriptstyle \pm 0.00}$ & 
$\textbf{0.76} {\scriptscriptstyle \pm 0.00}$ & 
$0.44 {\scriptscriptstyle \pm 0.01}$ \\ 
\hspace{3mm} + Bounding Box & & 
$\textbf{1.00} {\scriptscriptstyle \pm 0.00}$ & 
$0.43 {\scriptscriptstyle \pm 0.03}$ & 
$0.71 {\scriptscriptstyle \pm 0.00}$ & 
$0.34 {\scriptscriptstyle \pm 0.02}$ \\ 
\hspace{3mm} + Scene Graph & & 
$\textbf{1.00} {\scriptscriptstyle \pm 0.00}$ & 
$0.41 {\scriptscriptstyle \pm 0.00}$ & 
$0.71 {\scriptscriptstyle \pm 0.00}$ & 
$0.33 {\scriptscriptstyle \pm 0.02}$ \\ 
\hspace{3mm} + Depth Map & & 
$\textbf{1.00} {\scriptscriptstyle \pm 0.00}$ & 
$0.30 {\scriptscriptstyle \pm 0.01}$ & 
$0.69 {\scriptscriptstyle \pm 0.01}$ & 
$\textbf{0.46} {\scriptscriptstyle \pm 0.01}$ \\ 
\hspace{3mm} + Segmentation Map & & 
$\textbf{1.00} {\scriptscriptstyle \pm 0.00}$ & 
$0.28 {\scriptscriptstyle \pm 0.01}$ & 
$0.73 {\scriptscriptstyle \pm 0.02}$ & 
$0.45 {\scriptscriptstyle \pm 0.00}$ \\ 
\hspace{3mm} + Chain-of-Thought  & & 
$\textbf{1.00} {\scriptscriptstyle \pm 0.00}$ & 
$\textbf{0.63} {\scriptscriptstyle \pm 0.00}$ & 
$\textbf{0.76} {\scriptscriptstyle \pm 0.01}$ & 
$0.19 {\scriptscriptstyle \pm 0.1}$ \\ 
\hspace{3mm} + Rotation Grid & & 
$\textbf{1.00} {\scriptscriptstyle \pm 0.00}$ & 
$0.07 {\scriptscriptstyle \pm 0.01}$ & 
$0.57 {\scriptscriptstyle \pm 0.01}$ & 
$0.07 {\scriptscriptstyle \pm 0.01}$ \\ 
\hspace{3mm} + Rotation Grid Guided & & 
$\textbf{1.00} {\scriptscriptstyle \pm 0.00}$ & 
$0.10 {\scriptscriptstyle \pm 0.00}$ & 
$0.61 {\scriptscriptstyle \pm 0.00}$ & 
$0.25 {\scriptscriptstyle \pm 0.01}$ \\ 
\hspace{3mm} + \textit{all above} & & 
$\textbf{1.00} {\scriptscriptstyle \pm 0.00}$ & 
$0.1 {\scriptscriptstyle \pm 0.01}$ & 
$0.67 {\scriptscriptstyle \pm 0.01}$ & 
$0.17 {\scriptscriptstyle \pm 0.01}$ \\ 
\midrule
{\textit{GPT-4o}} \\ 
Zero-shot & & 
$0.99 {\scriptscriptstyle \pm 0.00}$ & 
$0.69 {\scriptscriptstyle \pm 0.02}$ & 
$0.93 {\scriptscriptstyle \pm 0.00}$ & 
$0.19 {\scriptscriptstyle \pm 0.01}$ \\ 
\hspace{3mm} + Caption & & 
$0.98 {\scriptscriptstyle \pm 0.00}$ & 
$0.65 {\scriptscriptstyle \pm 0.00}$ & 
$0.93 {\scriptscriptstyle \pm 0.01}$ & 
$0.23 {\scriptscriptstyle \pm 0.02}$ \\ 
\hspace{3mm} + Bounding Box & & 
$0.98 {\scriptscriptstyle \pm 0.01}$ & 
$0.59 {\scriptscriptstyle \pm 0.01}$ & 
$0.91 {\scriptscriptstyle \pm 0.00}$ & 
$0.31 {\scriptscriptstyle \pm 0.04}$ \\ 
\hspace{3mm} + Scene Graph & & 
$0.98 {\scriptscriptstyle \pm 0.00}$ & 
$0.55 {\scriptscriptstyle \pm 0.00}$ & 
$0.93 {\scriptscriptstyle \pm 0.00}$ & 
$0.33 {\scriptscriptstyle \pm 0.02}$ \\ 
\hspace{3mm} + Depth Map & & 
$\textbf{1.00} {\scriptscriptstyle \pm 0.00}$ & 
$0.55 {\scriptscriptstyle \pm 0.03}$ & 
$0.93 {\scriptscriptstyle \pm 0.00}$ & 
$0.26 {\scriptscriptstyle \pm 0.01}$ \\
\hspace{3mm} + Segmentation Map & & 
$0.97 {\scriptscriptstyle \pm 0.00}$ & 
$0.67 {\scriptscriptstyle \pm 0.01}$ & 
$\textbf{0.95} {\scriptscriptstyle \pm 0.00}$ & 
$0.21 {\scriptscriptstyle \pm 0.00}$ \\ 
\hspace{3mm} + Chain-of-Thought  & & 
$0.97 {\scriptscriptstyle \pm 0.01}$ & 
$0.57 {\scriptscriptstyle \pm 0.03}$ & 
$0.93 {\scriptscriptstyle \pm 0.00}$ & 
$0.32 {\scriptscriptstyle \pm 0.00}$ \\ 
\hspace{3mm} + Rotation Grid & & 
$0.98 {\scriptscriptstyle \pm 0.00}$ & 
$\textbf{0.71} {\scriptscriptstyle \pm 0.02}$ & 
$0.93 {\scriptscriptstyle \pm 0.00}$ & 
$0.19 {\scriptscriptstyle \pm 0.03}$ \\ 
\hspace{3mm} + Rotation Grid Guided & & 
$0.98 {\scriptscriptstyle \pm 0.00}$ & 
$0.46 {\scriptscriptstyle \pm 0.03}$ & 
$0.93 {\scriptscriptstyle \pm 0.00}$ & 
$\textbf{0.41} {\scriptscriptstyle \pm 0.00}$ \\ 
\hspace{3mm} + \textit{all above} & & 
$\textbf{1.00} {\scriptscriptstyle \pm 0.00}$ & 
$0.46 {\scriptscriptstyle \pm 0.03}$ & 
$0.91 {\scriptscriptstyle \pm 0.00}$ & 
$0.36 {\scriptscriptstyle \pm 0.02}$ \\ 
\bottomrule
\end{tabular*}
\caption{Classification accuracy using various forms of auxiliary information and prompting for all rotations. All results are obtained from three runs on \rotlarge{}. Accuracy is scored on a four-way classification task.
}
\label{tab:large_evals}
\end{table*}

\begin{table*}[ht]
\centering
\small 
\begin{tabular*}{0.85\textwidth}{@{\extracolsep{\fill}} llcccc}
\toprule
\multicolumn{2}{c}{} & \multicolumn{4}{c}{\textbf{Accuracy on Different Degrees of Rotation}} \\
\cmidrule(lr){3-6}
\multicolumn{2}{c}{\textbf{Model}} & \textbf{0°} & \textbf{90°} & \textbf{180°} & \textbf{270°} \\ 
\midrule
\textit{Human} \\
Zero-shot & & 
$0.99 $ & 
$0.99 $ & 
$0.99 $ & 
$0.97 $ \\   
\midrule
{\textit{\qwen{}}} \\
Zero-shot & & 
$\textbf{0.95} {\scriptscriptstyle \pm 0.01}$ & 
$\textbf{0.57} {\scriptscriptstyle \pm 0.04}$ & 
$0.03 {\scriptscriptstyle \pm 0.02}$ & 
$0.15 {\scriptscriptstyle \pm 0.05}$ \\ 
\hspace{3mm} + Chain-of-Thought  & & 
$0.86 {\scriptscriptstyle \pm 0.00}$ &
$0.20 {\scriptscriptstyle \pm 0.04}$ & 
$\textbf{0.13} {\scriptscriptstyle \pm 0.03}$ & 
$\textbf{0.30} {\scriptscriptstyle \pm 0.04}$ \\  
\midrule
{\textit{\gptfour{}}} \\ 
Zero-shot & & 
$0.87 {\scriptscriptstyle \pm 0.02}$ & 
$\textbf{0.65} {\scriptscriptstyle \pm 0.04}$ & 
$0.85 {\scriptscriptstyle \pm 0.01}$ & 
$\textbf{0.21} {\scriptscriptstyle \pm 0.03}$ \\ 
\hspace{3mm} + Chain-of-Thought  & & 
$\textbf{0.91} {\scriptscriptstyle \pm 0.01}$ & 
$0.59 {\scriptscriptstyle \pm 0.01}$ & 
$\textbf{0.86} {\scriptscriptstyle \pm 0.00}$ & 
$\textbf{0.21} {\scriptscriptstyle \pm 0.05}$ \\ 
\midrule
{\textit{\gptfourone{}}} \\ 
Zero-shot & & 
$0.95 {\scriptscriptstyle \pm 0.01}$ & 
$0.63 {\scriptscriptstyle \pm 0.07}$ & 
$\textbf{0.85} {\scriptscriptstyle \pm 0.03}$ & 
$\textbf{0.19} {\scriptscriptstyle \pm 0.09}$ \\ 
\hspace{3mm} + Chain-of-Thought  & & 
$\textbf{0.98} {\scriptscriptstyle \pm 0.00}$ & 
$\textbf{0.88} {\scriptscriptstyle \pm 0.00}$ & 
$\textbf{0.85} {\scriptscriptstyle \pm 0.01}$ & 
$0.03 {\scriptscriptstyle \pm 0.03}$ \\ 
\midrule
{\textit{\gptfive{}}} \\ 
Zero-shot & & 
$\textbf{1.00} {\scriptscriptstyle \pm 0.00}$ & 
$0.41 {\scriptscriptstyle \pm 0.03}$ & 
$\textbf{0.81} {\scriptscriptstyle \pm 0.01}$ & 
$\textbf{0.59} {\scriptscriptstyle \pm 0.04}$ \\ 
\hspace{3mm} + Chain-of-Thought  & & 
$\textbf{1.00} {\scriptscriptstyle \pm 0.00}$ & 
$\textbf{0.53} {\scriptscriptstyle \pm 0.05}$ & 
$\textbf{0.81} {\scriptscriptstyle \pm 0.02}$ & 
$0.57 {\scriptscriptstyle \pm 0.07}$ \\ 
\midrule
{\textit{\geminitwoflash{}}} \\ 
Zero-shot & & 
$\textbf{1.00} {\scriptscriptstyle \pm 0.00}$ & 
$0.25 {\scriptscriptstyle \pm 0.04}$ & 
$0.48 {\scriptscriptstyle \pm 0.01}$ & 
$\textbf{0.43} {\scriptscriptstyle \pm 0.07}$ \\ 
\hspace{3mm} + Chain-of-Thought  & & 
$\textbf{1.00} {\scriptscriptstyle \pm 0.00}$ & 
$\textbf{0.61} {\scriptscriptstyle \pm 0.01}$ & 
$\textbf{0.59} {\scriptscriptstyle \pm 0.01}$ & 
$0.25 {\scriptscriptstyle \pm 0.01}$ \\ 
\midrule
{\textit{\geminitwofiveflash{}}} \\ 
Zero-shot & & 
$\textbf{1.00} {\scriptscriptstyle \pm 0.00}$ & 
$\textbf{0.23} {\scriptscriptstyle \pm 0.06}$ & 
$\textbf{0.50} {\scriptscriptstyle \pm 0.03}$ & 
$\textbf{0.47} {\scriptscriptstyle \pm 0.02}$ \\ 
\hspace{3mm} + Chain-of-Thought  & & 
$\textbf{1.00} {\scriptscriptstyle \pm 0.00}$ & 
$\textbf{0.23} {\scriptscriptstyle \pm 0.01}$ & 
$0.44 {\scriptscriptstyle \pm 0.02}$ & 
$0.40 {\scriptscriptstyle \pm 0.02}$ \\ 
\midrule
{\textit{\othree{}}} \\ 
Zero-shot & & 
$\textbf{1.00} {\scriptscriptstyle \pm 0.00}$ & 
$\textbf{0.45} {\scriptscriptstyle \pm 0.04}$ & 
$0.70 {\scriptscriptstyle \pm 0.03}$ & 
$0.48 {\scriptscriptstyle \pm 0.03}$ \\ 
\hspace{3mm} + Chain-of-Thought  & & 
$\textbf{1.00} {\scriptscriptstyle \pm 0.00}$ & 
$0.36 {\scriptscriptstyle \pm 0.00}$ & 
$0.74 {\scriptscriptstyle \pm 0.04}$ & 
$0.57 {\scriptscriptstyle \pm 0.01}$ \\ 
\hspace{3mm} + Rotation Grid & & 
$0.99 {\scriptscriptstyle \pm 0.01} $ & 
$0.23 {\scriptscriptstyle \pm 0.05} $ & 
$\textbf{0.83} {\scriptscriptstyle \pm 0.05} $ & 
$\textbf{0.81} {\scriptscriptstyle \pm 0.01} $ \\ 
\hspace{3mm} + Rotation Grid Guided & &
$0.99 {\scriptscriptstyle \pm 0.01} $ & 
$0.31 {\scriptscriptstyle \pm 0.01} $ & 
$0.82 {\scriptscriptstyle \pm 0.02} $ & 
$0.75 {\scriptscriptstyle \pm 0.01} $ \\ 
\midrule
{\textit{\geminitwofivepro{}}} \\ 
Zero-shot & & 
$\textbf{1.00} {\scriptscriptstyle \pm 0.00}$ & 
$0.50 {\scriptscriptstyle \pm 0.04}$ & 
$0.72 {\scriptscriptstyle \pm 0.00}$ & 
$0.40 {\scriptscriptstyle \pm 0.06}$ \\ 
\hspace{3mm} + Chain-of-Thought  & & 
$\textbf{1.00} {\scriptscriptstyle \pm 0.00}$ & 
$0.46 {\scriptscriptstyle \pm 0.02}$ & 
$0.71 {\scriptscriptstyle \pm 0.01}$ & 
$0.49 {\scriptscriptstyle \pm 0.01}$ \\ 
\hspace{3mm} + Rotation Grid & &
$0.99 {\scriptscriptstyle \pm 0.01} $ & 
$0.58 {\scriptscriptstyle \pm 0.00} $ & 
$0.67 {\scriptscriptstyle \pm 0.03} $ & 
$0.59 {\scriptscriptstyle \pm 0.03} $ \\ 
\hspace{3mm} + Rotation Grid Guided & &
$0.95 {\scriptscriptstyle \pm 0.01} $ & 
$\textbf{0.71} {\scriptscriptstyle \pm 0.01} $ & 
$\textbf{0.73} {\scriptscriptstyle \pm 0.03} $ & 
$\textbf{0.74} {\scriptscriptstyle \pm 0.04} $ \\ 
\bottomrule
\end{tabular*}
\caption{Classification accuracy of different image rotations for various models across two runs on \rotsmall{} using zero-shot or chain-of-thought prompting. We also show results from \othree{} and \geminitwofivepro{} using rotation grids. Accuracy is scored on a four-way classification task.}
\label{tab:small_evals}
\end{table*}

\Cref{tab:large_evals} displays the results of evaluating \qwen{}, \geminitwoflash{}, and \gptfour{} along with various auxiliary information on \rotlarge{}. \Cref{tab:small_evals} displays results of evaluating \qwen{}, \gptfour{}, \gptfourone{}, \gptfive{}, \othree{}, \geminitwoflash{}, \geminitwofiveflash{}, \geminitwofivepro{} on \rotsmall{}. Due to the high token cost of proprietary reasoning models, we only evaluate zero-shot and CoT prompts. However, we evaluate providing rotation grids to \othree{} and \geminitwofivepro{}.\footnote{\Cref{tab:additional_results_reasoning} further shows results obtained from evaluating these two models on \rotsmall{} with all available auxiliary information.} \textcolor{black}{We provide qualitative examples of images that models answer correct and incorrectly in \cref{sec:qualitative_examples}.}

\paragraph{MLLMs accurately identify right-side-up (0°) images.} 
All evaluated models effectively recognize right-side-up (0°) images. \qwen{} (Qwen) achieves an accuracy of 0.99 without supplemental data. Proprietary models (\gptfour{}, \gptfourone{}, \othree{}, \geminitwofiveflash{}, \geminitwofivepro{}, and \geminitwoflash{}) consistently exhibit near-perfect accuracy on identifying unrotated images. This outcome aligns with expectations, given these models likely encountered predominantly upright images during training, and thus 0° can be assumed to be the \textit{default} option.

\paragraph{Proprietary models perform well on upside-down (180°) images.} 
All models except Qwen demonstrate robust performance on images rotated 180°, with \gptfour{}, \gptfourone{}, \othree{}, and \geminitwofivepro{} all achieving accuracies notably above chance (> 0.7). However, \geminitwoflash{} and \geminitwofiveflash{} display relatively lower zero-shot performance, with accuracies around 0.5. This indicates that state-of-the-art proprietary models generally possess a reliable capability to recognize upside-down images, though there remains variability within the different model families.

\paragraph{Identifying 90° and 270° is challenging for all evaluated models.}
All models exhibit substantial difficulties when distinguishing between 90° and 270°, while the poorest performance consistently emerged with 270° images. Confusion matrix analysis (\cref{sec:confusion_matrix}) reveals frequent misclassifications between these two orientations, indicating a distinct challenge when identifying 0° and 180°.

\paragraph{Providing auxiliary inputs does not reliably improve performance.}
None of the auxiliary information results in meaningful, consistent performance gains across all tested models. Paradoxically, the introduction of additional information sometimes leads to marginal performance degradation. When including all forms of auxiliary information, Qwen's accuracy on 90° decreased from 0.51 to 0.26. Similarly, \geminitwoflash{}'s accuracy on 270° decreased from 0.44 to 0.17. \textcolor{black}{We further examine why scene graphs fails to consistently improve model performance in \cref{sec:sg_ablation}.}

\paragraph{Rotation grid improves performance only for reasoning models.}
Providing the rotation grid degraded performance on most models. \geminitwoflash{} accuracy on 90° decreased by nearly 0.5 compared to CoT prompting. Rotation grid guided, however, improved \gptfour{} accuracy on 270° by around 0.1. At the same time, both \othree{} and \geminitwofivepro{} saw performance improvements. Notably, \geminitwofivepro{}'s accuracy on 90° and 270° improved by 0.15. These results suggest the two reasoning models are much more effective at utilizing visual context. We further leverage the robust identification of 0° images and modify the rotation grid into a majority voting approach (\Cref{sec:majority_vote_exp}). This setup results in gains even in weaker models, indicating MLLMs can achieve moderate performance with sufficient visual scaffolding.

\paragraph{CoT improves 180° performance, but results are mixed on 90° and 270°.}
Employing chain-of-thought (CoT) prompting yields mixed results. Gemini models show improved accuracy on 90° rotations but decreased accuracy for 270°. On the other hand, GPT models displayed inverse trends. Yet, CoT consistently enhanced accuracy for 180° rotations across all models tested. These findings suggest CoT prompting can help models better reason about rotations to some extent, but does not universally resolve inherent challenges in distinguishing portrait orientations. \Cref{sec:cot_trace} closely examines a common failure of CoT.
 
\section{Additional Analyses}

\subsection{Model Bias Towards 0° and 90°}
\label{sec:confusion_matrix}

\begin{figure}[ht]
    \centering
    \small
    \includegraphics[width=0.35\textwidth]{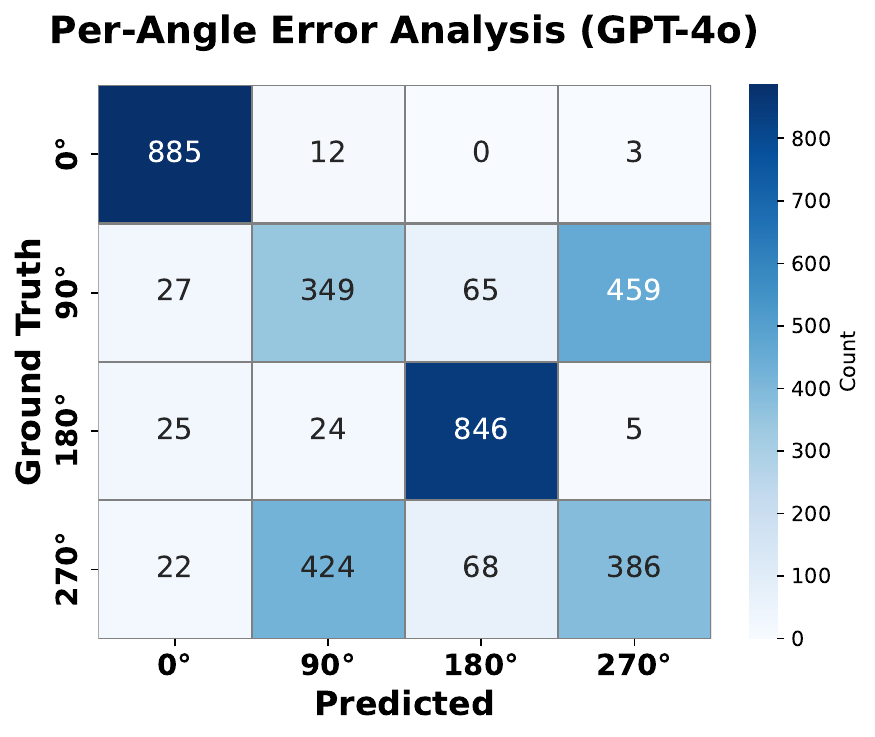}
    \caption{Confusion matrix of true vs. predicted rotations for \gptfour{} using CoT prompting, summed across three runs on \rotlarge{}. Rows represent ground-truth labels, columns represent predicted labels. The matrix highlights a significant confusion specifically between 90° and 270° rotations.
    }
    \label{tab:gpt_4o_confusion_cot}
\end{figure}

To further elucidate the specific types of rotational errors made by models, we analyze the confusion matrix for \gptfour{}. \Cref{tab:gpt_4o_confusion_cot} shows the confusion matrix obtained from summing predictions across three runs. We see \gptfour{} predominantly struggles with differentiating between 90° and 270° rotations. Specifically, the model misclassifies 459 instances of 90° images as 270° and 424 instances of 270° images as 90°. Yet, 0° and 180° show significantly fewer misclassifications. This analysis underscores a critical shortcoming in model performance, suggesting either that the vision encoder is not providing sufficient signals to distinguish between clockwise and counter-clockwise rotations, or that the MLLM is not adequately incorporating the visual information into its reasoning. 

\subsection{Distinguishing Clockwise from Counter-clockwise Rotations}
\label{sec:cw_ccw_experiment}

\begin{table}[t]
\centering
\small
\begin{tabular}{lSS}
\toprule
\textbf{GT} $\backslash$ \textbf{Predicted} &
\textbf{Counter-clockwise} & \textbf{Clockwise} \\
\midrule
\multicolumn{3}{l}{\textit{GPT-4o}} \\
\cmidrule(lr){1-3}
Counter-clockwise & 248 & 52 \\
Clockwise         & 259 & 41 \\
\midrule
\multicolumn{3}{l}{\textit{Qwen-2.5-VL-7B-Instruct}} \\
\cmidrule(lr){1-3}
Counter-clockwise & 270 & 30 \\
Clockwise         & 277 & 23 \\
\bottomrule
\end{tabular}
\caption{Accuracy of different models in identifying 90° clockwise (CW) versus counter-clockwise rotation (CCW). Column indicates ground truth rotation (GT). A small number of responses are incorrectly identified as being right-side up. These responses are excluded from the table's statistics.
}
\label{tab:cww_test}
\end{table}

\myparagraph{Setup.}
\begin{figure}[ht]
    \centering
    \small
    \includegraphics[width=0.4\textwidth]{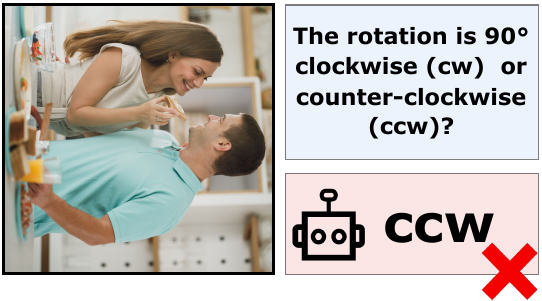}
    \caption{\gptfour{} answers incorrectly when asked to identify whether the image has been rotated 90° clockwise or counter-clockwise.
    }
    \label{fig:img_cw_ccw}
\end{figure}
Following the analysis in \cref{sec:confusion_matrix}, we examine whether MLLMs can distinguish between clockwise (CW) and counter-clockwise rotations (CCW) by simplifying the four-way classification task into a binary classification task. Using the 90° and 270° images from \rotlarge{}, we prompt MLLMs to determine whether an image has been rotated 90° CCW or CW.\footnote{Note that 90° CW is equivalent to a 270° CCW.} By asking models to explicitly differentiate the two directions, we hope to provide a clearer signal for directional understanding. \Cref{fig:img_cw_ccw} shows an example question where the \gptfour{} provides an incorrect answer.

\myparagraph{Results.}
\Cref{tab:cww_test} tests CW vs. CCW classification on \gptfour{} and \qwen{} (we additionally provide similar smaller-scale tests on \rotsmall{} for \gptfive{} in \cref{sec:gpt5_cw_ccw}).
\cref{tab:cww_test} indicates that \gptfour{} has a significant bias toward identifying rotations as 90° counter-clockwise. 
\gptfour{} correctly identified only 52 out of 300 clockwise rotations, whereas \qwen{} correctly identified only 23 out of 300 clockwise rotations. The models consistently default to labeling ambiguous or uncertain rotations as counter-clockwise, hinting towards a potential underlying perceptual bias (\cref{sec:cot_trace}).
While we find that the most recent model tested, \gptfive{}, has less bias and is better able to identify clockwise rotations, this improvement is not reflected in \cref{tab:small_evals}, where the model performs on par with others; in other words, poor performance on \rotbench{} cannot be solely explained by a model's ability to distinguish clockwise from counterclockwise rotation. See \cref{sec:gpt5_cw_ccw} for further discussion. These findings strongly indicate that the MLLMs tested have limitations in reliably distinguishing between CW and CCW rotational directions, offering a potential explanation for why distinguishing 90° and 270° CCW rotations presents such a challenging task. Further corroborating this idea, we show that evaluating \qwen{}, \gptfour{}, \othree{} on \rotsmall{} using clockwise angles does not result in significant performance differences (\cref{sec:cw_prompting}).

\subsection{Can Fine-tuning Solve \rotbench{}?}
\label{sec:ft_experiment}

\begin{figure}[htbp]
    \centering
    \small
    \includegraphics[width=0.48\textwidth]{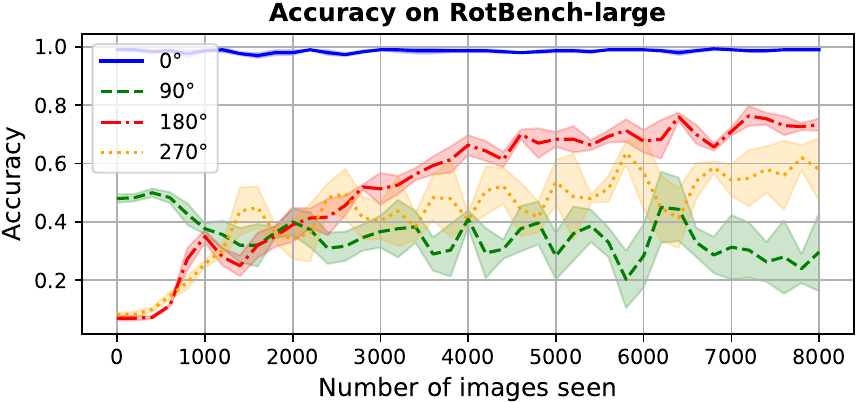}
    \caption{\qwen{}'s accuracy on different degrees of rotation as training progresses.}
    \label{fig:ft_exp}
\end{figure}

\myparagraph{Setup.}
To assess whether specialized training can mitigate these performance issues, we conduct fine-tuning experiments using \qwen{}. During data filtering, we find some images in Spatial-MM closely resemble each other (\cref{fig:similar_imgs}). If we fine-tune on images in Spatial-MM not selected for \rotbench{}, the existence of similar images in the training and testing sets may lead to inflated performance. To prevent such instances of overfitting, we train on 1000 images from MS COCO \cite{lin2015coco} and evaluate performance on \rotlarge{}. \Cref{sec:computation_details} provides further training details. 

\myparagraph{Results.}
\cref{fig:ft_exp} reveals a high and consistent accuracy for 0° throughout training, indicating robust recognition of upright images. Performance on 180° gradually improves, stabilizing around 0.8 after approximately 7000 images. However, accuracies for 90° and 270° exhibit substantial oscillations, suggesting that the model did not achieve stable improvement. The model appears to be caught in a cycle, alternating between accuracy gains and losses for these two rotations. This phenomenon is also reflected in our main results (\cref{sec:main_results}), where using CoT prompting improved accuracy on 270° images at the expense of 90°. The unstable performance may result from potential representational constraints in current visual encoders that limit visual understanding capabilities, particularly regarding subtle rotational distinctions.

\subsection{Normalized Rotation Voting} 
\label{sec:majority_vote_exp}

\begin{table}[ht]
\centering
\small 
\begin{tabular}{lcccc}
\toprule
\multicolumn{1}{c}{} & \multicolumn{4}{c}{\textbf{Accuracy (↑)}} \\
\cmidrule(lr){2-5}
\multicolumn{1}{c}{\textbf{Model}} & \textbf{0°} & \textbf{90°} & \textbf{180°} & \textbf{270°} \\ 
\midrule
{\textit{\qwen{}}} \\
Zero-shot & 
$\textbf{0.95} $ & 
$\textbf{0.57} $ & 
$0.03 $ & 
$0.15 $ \\ 
Chain-of-Thought  &
$0.86 $ & 
$0.20 $ & 
$0.13 $ & 
$0.30 $ \\ 
Normalized Rotation Voting &
$0.54 $ & 
$0.52 $ & 
$\textbf{0.56} $ & 
$\textbf{0.52} $ \\ 
\midrule
{\textit{\gptfour{}}} \\ 
Zero-shot & 
$0.87 $ & 
$0.65 $ & 
$0.85 $ & 
$0.21 $ \\ 
Chain-of-Thought  &
$\textbf{0.91} $ & 
$0.59 $ & 
$\textbf{0.86} $ & 
$0.21 $ \\ 
Normalized Rotation Voting &
$0.86 $ & 
$\textbf{0.88} $ & 
$0.80 $ & 
$\textbf{0.86} $ \\ 
\bottomrule
\end{tabular}
\caption{Average classification accuracy under different image rotation angles (0°, 90°, 180°, 270°) using normalized rotation voting on \rotsmall{}. Accuracy is scored on a four-way classification task. 
}
\label{tab:voting_results}
\end{table}

\begin{algorithm}
\caption{Normalized Rotation Voting}
\label{alg:rotation_vote}
\begin{algorithmic}[1]
  \Require $images = [I_0, I_{90}, I_{180}, I_{270}]$
  \State $rotations \gets []$
  \For{$i \gets 0$ to $|images|-1$}
    \State $rot \gets \text{call\_model}(images[i])$
    \State $rot\_norm \gets (rot - i \times 90)\bmod 360$
    \State append $rot\_norm$ to $rotations$
  \EndFor
  \State $final\_rot \gets \text{majority\_vote}(rotations)$
\end{algorithmic}
\end{algorithm}

\myparagraph{Intuition and Setup.}
Our results show frontier MLLMs are able to reliably identify 0° and 180° images. Leveraging this pattern, we propose a voting approach to identify image rotation that explicitly exploits this asymmetry in model behavior. Given an input image of unknown orientation, we can \textit{further} rotate this image 0°, 90°, 180°, and 270°. We separately prompt the model to identify the rotation of each image, then normalize the predictions by subtracting the applied rotation, effectively shifting the angle into a common reference frame (\cref{alg:rotation_vote}). Regardless of the original rotation, two of the further rotated images would correspond to 0° and 180° in the ground-truth reference frame. If the model can correctly identify these two images, there is a high probability that the majority vote would reveal the ground-truth rotation. We evaluate \gptfour{} and \qwen{} on \rotsmall{} using this normalized majority voting approach to test whether our intuition hold empirically.

\myparagraph{Results.}
\Cref{tab:voting_results} shows the accuracy on each image orientation using normalized rotation voting. In general, we see a much more even performance distribution across all four rotations compared to using zero-shot or chain-of-thought prompting. \qwen{} achieves substantially stronger zero-shot performance on 0° and 90° (0.95 and 0.57) compared to 180° and 270° (0.03 and 0.15). Normalized rotation voting achieves $0.5$ performance on all orientations. \gptfour{} achieves $0.8$ or above accuracy on all orientations. In particular, performance on 270° improved substantially compared to zero-shot prompting.

These results show that this voting approach provides a viable means of identifying image rotation using MLLMs. However, this approach has significant drawbacks. Firstly, it increases the compute required to predict a single image, as the model processes four images for each input image. Secondly, it relies on the assumption that we have prior knowledge of all possible rotations of the image. In real-world use cases, the rotation of an input image is most likely a continuous value. While it is possible to discretize continuous rotations to a level of granularity, this approach would quickly become impractical as finer granularity increases model calls.

\section{Conclusion}

We evaluate whether MLLMs are able to identify input images rotated 0°, 90°, 180°, and 270° using \rotbench{}, a 350-image manually-filtered benchmark. Our results reveal that state-of-the-art MLLMs reliably identify images that are unrotated (0°) or upside-down (180°), but struggle with 90° and 270° rotations. Auxiliary information and chain-of-thought prompting provide limited improvements. Simultaneously providing all possible rotations improves performance only for frontier reasoning models, but a modified setup using majority voting improves performance for weaker models as well. Fine-tuning \qwen{} shows that performance oscillates between 90° and 270°, suggesting the presence of two local optima. These results indicate a potential blind spot in MLLMs' spatial reasoning capabilities, motivating future directions in improving orientational understanding. 

\section*{Limitations}

Our work shows MLLMs struggle to accurately identify the rotation of input images rotated (0°, 90°, 180°, and 270°). In real-world scenarios, images are typically rotated by arbitrary angles. Therefore, our work may in fact overestimate model capabilities. Moreover, we did not evaluate larger open-weight models due to resource and hardware limitations. Rather, we evaluated the most recent and largest proprietary MLLMs through their APIs. Since these proprietary models dominate open-weight models on nearly all benchmarks \citep{openai2025o3, gemini2025gemini2.5}, we expect larger open-weight models to perform worse or on par with the top proprietary models tested. 

\section*{Acknowledgements}
This work was supported by DARPA ECOLE Program No. HR00112390060,
NSF-CAREER Award 1846185, NSF-AI Engage Institute DRL-2112635, ARO Award W911NF2110220, ONR Grant N00014-23-1-2356, and a Bloomberg Data Science PhD Fellowship. 
The views contained in this article are those of the authors and not of the funding agency.

\bibliography{custom}

\appendix

\section*{Appendix}

\section{Dataset Details}
\label{sec:data_details}

\subsection{Further details on Stage 1 and 2 Filtering}
\begin{figure*}[htbp]
    \centering
    \includegraphics[width=\textwidth]{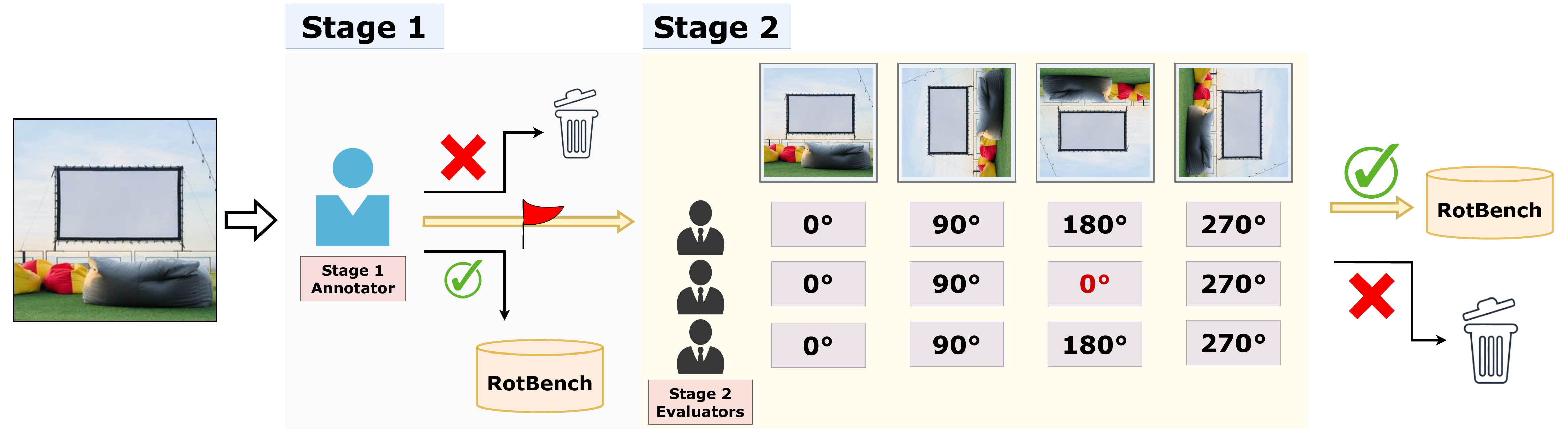}
    \caption{Figure describes our two-stage data filtering procedure. The example image is flagged during Stage 1, but subsequently accepted during Stage 2 as only one evaluator provided an incorrect response.}
    \label{fig:data_filtering}
\end{figure*}

\begin{figure*}[htbp]
    \centering
    \includegraphics[width=0.8\textwidth]{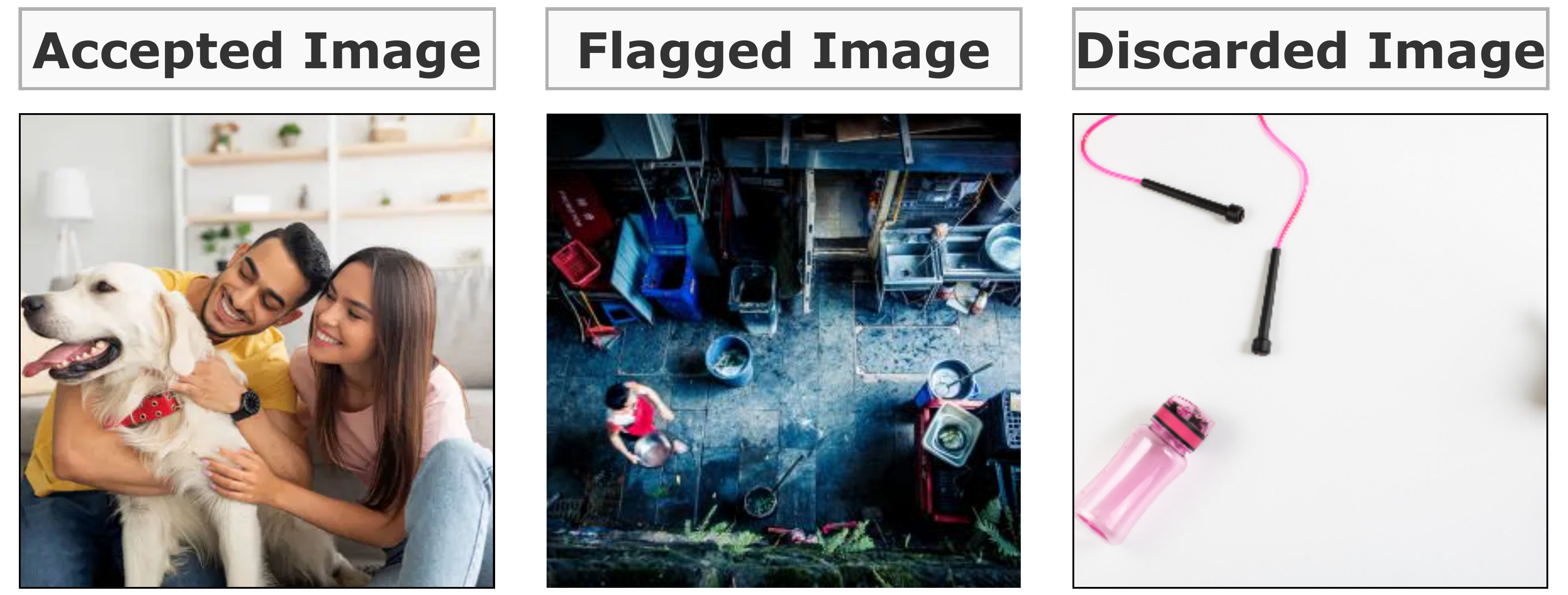}
    \caption{Samples of accepted, flagged, and discarded images during Stage 1 of data filtering.}
    \label{fig:appendix_sample_imgs}
\end{figure*}

\myparagraph{Stage 1.}
We randomly sample 300 images from Spatial-MM \cite{shiri2024empirical}. Spatial-MM is divided into two splits, \textit{one-subject} and \textit{two-subject}, respectively comprised of images that contain one and two primary subjects. As models may use image dimensions to infer orientation, we crop each image into a square before further processing. We sample 150 images from each split and show the resulting 300-image dataset to the Stage 1 annotator. For each image, the annotator can decide to accept, discard, or flag. The annotator accepts an image if it (1) contains easily identifiable rotational visual cues (e.g., a person standing), and (2) the image has meaningful differences when rotated. Images that do not satisfy these criteria are discarded. Occasionally, some images have subtle visual cues, or require more detailed semantic understanding to correctly perceive orientation. These images tend to have primary subjects that do not vary significantly with rotation, or require integrating background signals to identify rotation. The sample image shown in \cref{fig:data_filtering} is an example of a flagged image. Such images are flagged by the Stage 1 annotator to proceed to Stage 2. We provide further examples and details of accepted, flagged, and discarded images in \cref{fig:appendix_sample_imgs}. Note that the Stage 1 annotator is also tasked with ensuring no leaked personal information or offensive content is incorporated into \rotbench{}. 

\myparagraph{Stage 2.}
Stage 2 involves a group of three human evaluators. Each flagged image is rotated 0°, 90°, 180°, and 270°, producing four images. We shuffle all images -- which now include rotated images -- and present them to the evaluators as multiple-choice questions (\cref{sec:annotator_interface}). Any image that elicits an incorrect answer from two or more evaluators across all four orientations is discarded. Otherwise, the image is accepted.

Of the first 300 sampled images, only 27 were flagged. Among the flagged images, only two were eventually rejected. We sample two additional images to replace the discarded images. These images form the 300-image \rotlarge{}.

\subsection{Creating \rotsmall}

As each image is rotated in four orientations (0°, 90°, 180°, and 270°), obtaining human performance on \rotlarge{} is costly. 
We therefore propose \rotsmall{}, a human-evaluated 50-image subset where we can establish a human baseline. Starting from the 25 non-discarded flagged images in Stage 2, we obtain \rotsmall{} by further sampling 25 more images from Spatial-MM that fit the criteria for flagging. We ensure an equal number of one and two primary subject images in the final 50-image dataset. We then repeat the Stage 2 procedure on this new set, obtaining a human baseline. Notably, none of the 25 new images were discarded from human evaluation.

\subsection{Annotator Information}
\label{sec:annotator_info}

All three evaluators of Stage 2 data filtering are university students majoring in Computer Science, and have prior experience in Artificial Intelligence research. Two evaluators are undergraduate students who have volunteered to participate in the project, the third is a graduate student who also provided Stage 1 annotations and is an author of this work. All evaluators consented to have their data incorporated.

\begin{figure}[htbp]
    \centering
    \includegraphics[width=.5\textwidth]{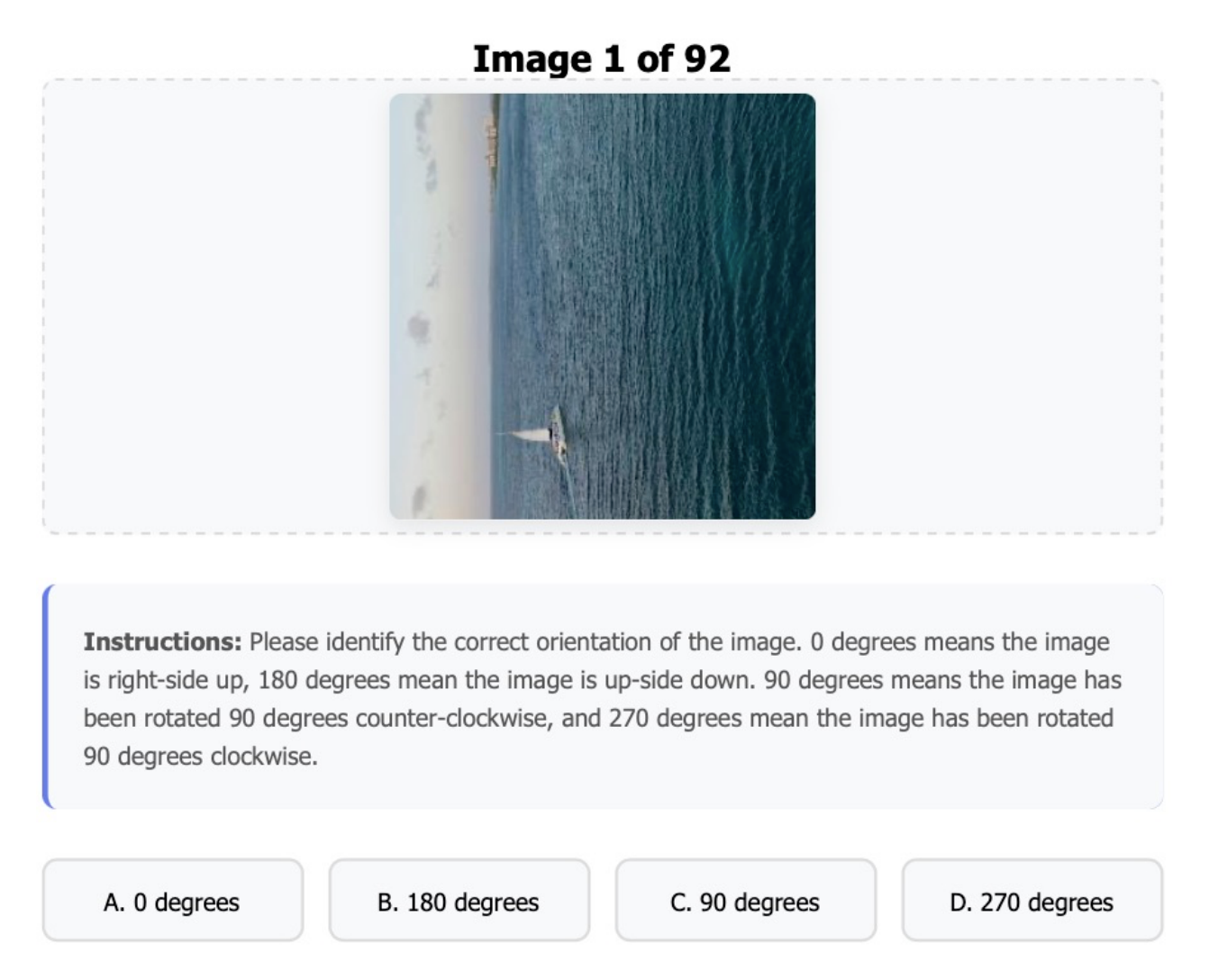}
    \caption{A screenshot of the custom interface shown to Stage 2 annotators.}
    \label{fig:data_interface}
\end{figure}

\begin{figure*}[htbp]
    \centering
    \includegraphics[width=0.88\textwidth]{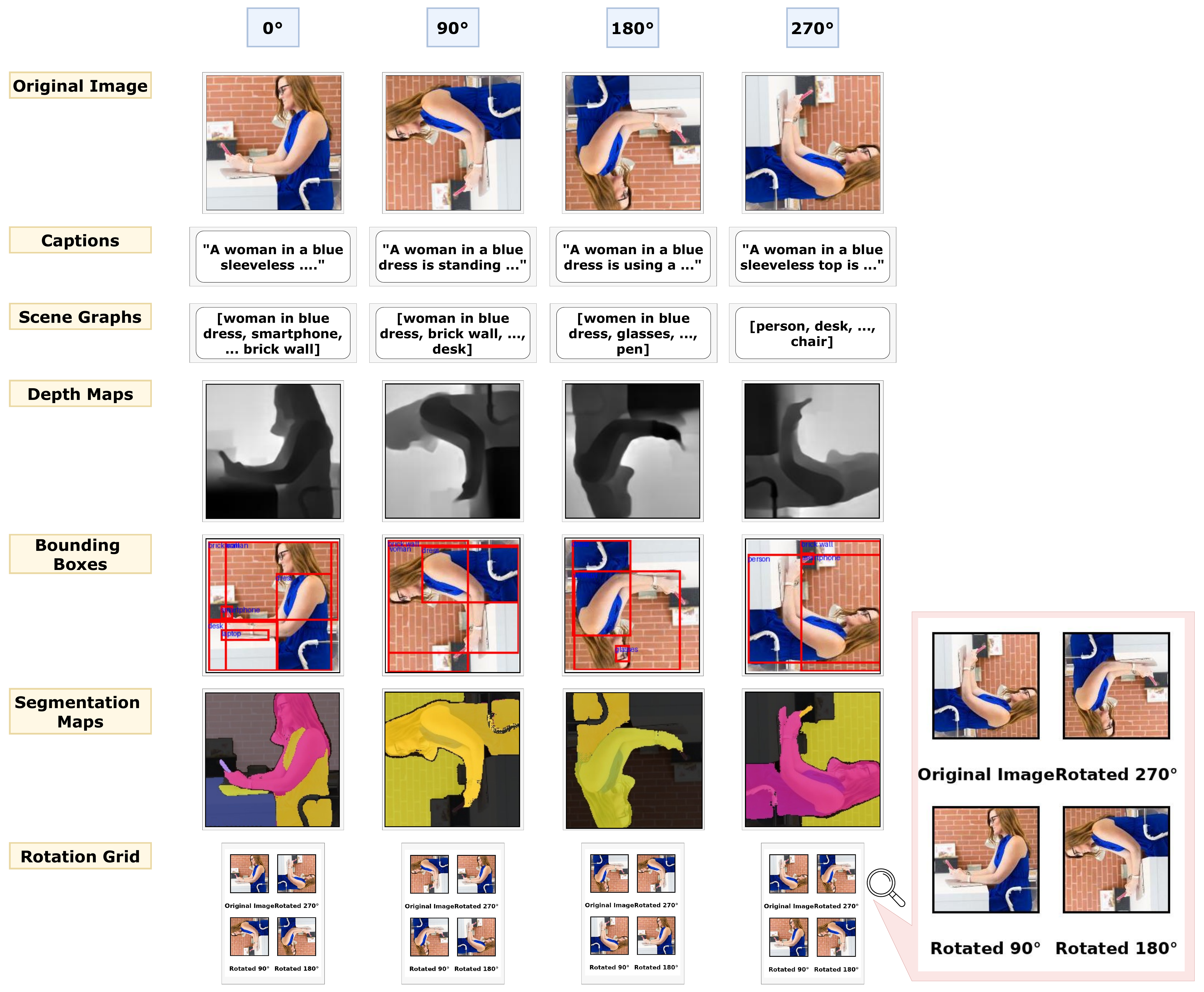}
    \caption{Examples of the different types of auxiliary information provided to the models.}
    \label{fig:appendix_aux_info}
\end{figure*}

\subsection{Annotator Interface}
\label{sec:annotator_interface}

\Cref{fig:data_interface} shows the interface provided to Stage 2 annotators. The mapping between letter choice and rotation degree is shuffled from question to question, ensuring the model does not suffer from choice-order bias.

\subsection{Sample Images}
\label{sec:sample_images}

\cref{fig:appendix_aux_info} displays the various forms of additional information provided to the model. \cref{fig:appendix_sample_imgs} provides examples of an image that is accepted, flagged, and discarded during Stage 1 of data filtering. The accepted image has a clear subject and is easily distinguishable when rotated 0°, 90°, 180°, and 270°. The flagged image is more difficult to identify; however, the slight tilt---as opposed to a directly top-down perspective---still enables accurate judgment of rotation. On the other hand, the discarded image has no meaningful signals to distinguish between the four orientations.

\subsection{Further details on flagged images}

Flagged images typically involve cases where there is a lack of primary subjects or the primary subject does not offer any distinguishable signals under rotation. These images tend to require deeper comprehension to achieve accurate classification. For instance, \cref{fig:data_filtering} displays a blank projector canvas above some colorful couches on a patio. The orientation of a projector screen cannot be reliably identified by where its supporting structures are located---the screen can be hung (cable hangs down), pitched up (supporting stand on the ground), or attached to a wall (support frame located sideways of the screen). Therefore, correctly identifying the orientation of the image requires understanding the correct orientation of the couches.

\subsection{\textcolor{black}{Motivation for Choosing Spatial-MM}}

There are multiple ways a model may reason about rotation. If an image has only one primary subject, rotation must be identified using the subject itself and background information. However, if there are multiple primary subjects, the spatial relationship between the two can also provide valuable information. We wish to curate a dataset that balances these two forms of reasoning. As previously mentioned, Spatial-MM’s images already distinguish between these two categories, greatly facilitating this balancing when creating \rotbench{}. Furthermore, Spatial-MM is composed of images scraped from the internet, covering a wide range of lifestyle, portrait, and landscape images. This diversity leads to better generalization to other real-world scenarios.

\section{\textcolor{black}{Other Relevant Work}}
\label{sec:further_relevant_work}

\textcolor{black}{
\myparagraph{Camera orientation estimation.}
Camera orientation estimation is also a well-studied task in computer vision \cite{xian2019uprightnet}. Instead of predicting the rotation of the image, camera orientation estimation seeks to predict the spatial location of the camera when capturing an image. Contemporary approaches of this task use deep networks to directly predict orientation parameters from image features in an end-to-end manner \cite{xian2019uprightnet, lee2021ctrlc, lee2020neuralgeometricparser}.
}

\myparagraph{Sensitivity to visual perturbations.}
Previous work has shown that visual encoders and MLLMs are sensitive to simple image transformations. \citet{anis2025} evaluate CLIP \cite{radford2021clip} and SigLIP \cite{zhai2023sigclip} on a suite of common image transformations---rotations, flips, noise, etc.---revealing substantial gaps between human and model understanding. \citet{usama2025} finds MLLMs exhibit distinct failure patterns in scene‐text and object reasoning tasks when applying ImageNet-C corruptions \cite{hendrycks2019imagenetc} to image inputs. Together, these studies highlight that despite strong clean‐image performance, visual encoders and MLLMs are highly sensitive to photometric and geometric distortions. 

\myparagraph{Robustness to image transformations.}
Past work has also examined various methods to ensure that image transformations do not affect downstream task performance. \citet{agnieszka2018}, \citet{shorten2019}, and \citet{perez2017} use image transformations as data augmentation methods to improve downstream classifier robustness. Other works instead proposed alternative architectures and training schemes to improve robustness to rotation \cite{xu2023e2, cohen2016group, lee2023learning, feng2019selfsupervised}. While this line of work focuses on training models to ignore certain transformations and learn invariant features for downstream tasks, we instead focus on identifying and reasoning about the transformation itself. 

\myparagraph{Identifying visual perturbations.}
Past work has tested how well vision encoders and MLLMs can identify which perturbation or transformation was applied to an input image \citep{lin2020visual, an2023more, rashtchian2023substance}. \citet{udandarao2024visual} show that visual encoders and MLLMs often fail to classify which perturbation out of 27 perturbations (e.g., contrast, brightness, rotation, blur) was applied to an input image on automatically-created datasets. In contrast, we provide a fine-grained analysis of rotation specifically through a 4-way angle classification task on a curated dataset with rigorous human verification, offering novel insights into why state-of-the-art modern MLLMs struggle with distinguishing specific rotation angles. Moreover, we test providing remedies such as chain-of-thought reasoning, various forms of auxiliary information (e.g., captions, bounding boxes, segmentation maps, depth maps), and guiding reasoning with algebraic calculations. We also analyze whether model performance is dependent on abilities such as identifying clockwise versus counter-clockwise rotation.

\section{\textcolor{black}{Generalization and Potential Downstream Applications}}
\label{sec:downstream_apps}

While our work provides a detailed analysis into the task of identifying image rotation, showing that such a simple and intuitive task for humans remains challenging for frontier reasoning MLLMs. Despite its straightforward nature, image rotation estimation requires strong perception, spatial reasoning abilities, and general world knowledge. For each image, the model must recognize which objects are relevant to rotational understanding, identify their spatial relationships, and determine whether there exists any irregularities in such relationships. Furthermore, computer-vision tasks typically tend to emphasize the primary subject, or subjects at the forefront of an image (e.g., object extraction, counting, segmentation, etc). However, image rotation often requires the model to also interpret subtle background information.

These abilities are a fundamental prerequisite for higher-level spatial reasoning. Our findings expose systematic blind spots which can lead to failures in more complex downstream tasks, such as understanding relative positions (“left/right/above/below”) or reasoning under rotated cameras.

In addition, we further elaborate on the two downstream applications outlined in \cref{sec:intro}:

\begin{itemize}
  \item Robotics: AI tasks in manufacturing may require cameras mounted on robotic arms that can move in multiple axes. Researchers interested in training a robotic arm to perform complex maneuvers – such as orienting a drill bit to reach awkward angles – with a high degree of autonomy, would require a model that can accurately identify the current rotation of the arm.
  \item First-person Point-of-View (PoV) footage analysis: Analyzing footage from various extreme sports – such as cliff diving or sky diving – and aerial vehicles – such as camera drones – involves complex rotational movement across multiple axes of rotation. Researchers interested in training RL agents in such environments require the agent to exhibit strong spatial reasoning abilities, among which is the prerequisite ability to identify its current rotation from the image.
\end{itemize}

\section{Model Inference and Training Details}
\label{sec:computation_details}

All proprietary models are accessed through first-party APIs provided in their respective official Python packages; the open-weight models are run on an Nvidia RTX A6000 Ada GPU (48 GB). When fine-tuning \qwen{}, we employ 4-bit quantization with the Bits-And-Bytes (bnb) NF4 format \cite{dettmers2023qlora} and apply Low-Rank Adaptation (LoRA) \cite{hu2021loralowrankadaptationlarge} with a rank 8 and an alpha parameter of 32. The fine-tuning procedure ran for 2 epochs, with a batch size of 32 and a learning rate of 2e-5.

\section{Additional Results}
\label{sec:additional_results}

We provide experimental results from evaluating \llama{} on \rotlarge{} (\cref{tab:additional_results_llama}), as well as \othree{}, \geminitwofivepro{} \textcolor{black}{(\cref{tab:additional_results_reasoning}),\smallqwen{}, \claude{}, \gglphi{}, \gemma{}, and \molmo{} on \rotsmall{} (\cref{tab:additional_results_other_models}}. \llama{} seems to be the weakest model tested, showing inferior performance across all rotations. Without CoT, the model only achieves roughly 0.3 accuracy. The two frontier reasoning models (\othree{} and \geminitwofivepro{}) show no significant performance improvement when prompted with auxiliary information. However, they respond positively to rotation grids. \geminitwofivepro{} achieves the highest accuracy on 90° and 270° (both 0.7) using guided rotation grids, while \othree{} achieves its highest accuracy on 180° (0.83) using the rotation grid. These results indicate that strong reasoning models are able to utilize the grid to aid them in making a prediction, while weaker models struggle with it. 

\begin{table*}[ht]
\centering
\small 
\begin{tabular*}{0.75\textwidth}{@{\extracolsep{\fill}} llcccc}
\toprule
\multicolumn{2}{c}{} & \multicolumn{4}{c}{\textbf{Accuracy on Different Degrees of Rotation}} \\
\cmidrule(lr){3-6}
\multicolumn{2}{c}{\textbf{Model}} & \textbf{0°} & \textbf{90°} & \textbf{180°} & \textbf{270°} \\ 
\midrule
{\textit{Llama-3.2-11B-Instruct}} \\ 
Zero-shot & & 
$0.28 {\scriptscriptstyle \pm 0.01}$ & 
$0.14 {\scriptscriptstyle \pm 0.02}$ & 
$0.53 {\scriptscriptstyle \pm 0.01}$ & 
$0.30 {\scriptscriptstyle \pm 0.02}$ \\ 
\hspace{3mm} + Caption & & 
$0.36 {\scriptscriptstyle \pm 0.03}$ & 
$0.11 {\scriptscriptstyle \pm 0.02}$ & 
$\textbf{0.64} {\scriptscriptstyle \pm 0.02}$ & 
$0.25 {\scriptscriptstyle \pm 0.03}$ \\ 
\hspace{3mm} + Bounding Box & & 
$0.22 {\scriptscriptstyle \pm 0.01}$ & 
$0.20 {\scriptscriptstyle \pm 0.02}$ & 
$0.43 {\scriptscriptstyle \pm 0.01}$ & 
$0.26 {\scriptscriptstyle \pm 0.02}$ \\ 
\hspace{3mm} + Scene Graph & & 
$0.27 {\scriptscriptstyle \pm 0.01}$ & 
$0.17 {\scriptscriptstyle \pm 0.02}$ & 
$0.43 {\scriptscriptstyle \pm 0.02}$ &
$0.26 {\scriptscriptstyle \pm 0.01}$ \\ 
\hspace{3mm} + Chain-of-Thought  & & 
$0.45 {\scriptscriptstyle \pm 0.01}$ & 
$\textbf{0.28} {\scriptscriptstyle \pm 0.02}$ & 
$0.31 {\scriptscriptstyle \pm 0.01}$ & 
$0.34 {\scriptscriptstyle \pm 0.00}$ \\ 
\hspace{3mm} + \textit{all above} & & 
$\textbf{0.47} {\scriptscriptstyle \pm 0.04}$ &
$0.16 {\scriptscriptstyle \pm 0.01}$ & 
$0.41 {\scriptscriptstyle \pm 0.02}$ & 
$\textbf{0.40} {\scriptscriptstyle \pm 0.00}$ \\ 
\bottomrule
\end{tabular*}
\caption{Average classification accuracy under different image rotation angles (0°, 90°, 180°, 270°) for \llama{} and auxiliary information across three runs on \rotlarge{}. Accuracy is scored on a four-way classification task. We did not evaluate providing depth maps or segmentation maps for \llama{} (Llama) as it only supports a single image input.
}
\label{tab:additional_results_llama}
\end{table*}

\begin{table}[ht]
\centering
\small 
\begin{tabular}{lcccc}
\toprule
\multicolumn{1}{c}{} & \multicolumn{4}{c}{\textbf{Accuracy (↑)}} \\
\cmidrule(lr){2-5}
\multicolumn{1}{c}{\textbf{Model}} & \textbf{0°} & \textbf{90°} & \textbf{180°} & \textbf{270°} \\ 
\midrule
{\textit{\geminitwofivepro{}}} \\
Zero-shot & 
$\textbf{1.00} $ & 
$0.50 $ & 
$0.72 $ & 
$0.40 $ \\ 
\hspace{3mm} + Caption &
$\textbf{1.00} $ & 
$0.48 $ & 
$0.76 $ & 
$0.44 $ \\ 
\hspace{3mm} + Bounding Box &
$0.98$ & 
$0.60 $ & 
$0.74 $ & 
$0.34 $ \\ 
\hspace{3mm} + Scene Graph & 
$\textbf{1.00} $ & 
$0.54 $ & 
$\textbf{0.78} $ & 
$0.42 $ \\ 
\hspace{3mm} + Depth Map &
$\textbf{1.00} $ & 
$0.44 $ & 
$0.66 $ & 
$0.52 $ \\ 
\hspace{3mm} + Segmentation Map &
$0.98$ & 
$0.54 $ & 
$0.70 $ & 
$0.42 $ \\ 
\hspace{3mm} + Chain-of-Thought  &
$\textbf{1.00} $ & 
$0.23 $ & 
$0.44 $ & 
$0.40 $ \\ 
\hspace{3mm} + Rotation Grid &
$0.99$ & 
$0.58 $ & 
$0.67 $ & 
$0.59 $ \\ 
\hspace{3mm} + Rotation Grid Guided &
$0.95$ & 
$\textbf{0.71} $ & 
$0.73 $ & 
$\textbf{0.74} $ \\ 
\hspace{3mm} + \textit{all above} &
$\textbf{1.00} $ & 
$0.60 $ & 
$0.76 $ & 
$0.68 $ \\ 
\midrule
{\textit{\othree{}}} \\ 
Zero-shot & 
$\textbf{1.00} $ & 
$\textbf{0.45} $ & 
$0.70 $ & 
$0.48 $ \\ 
\hspace{3mm} + Caption &
$\textbf{1.00} $ & 
$0.32 $ & 
$0.70 $ & 
$0.44 $ \\ 
\hspace{3mm} + Bounding Box &
$\textbf{1.00} $ & 
$0.38 $ & 
$0.64 $ & 
$0.40 $ \\ 
\hspace{3mm} + Scene Graph &
$\textbf{1.00} $ & 
$0.40 $ & 
$0.60 $ & 
$0.48 $ \\ 
\hspace{3mm} + Depth Map &
$\textbf{1.00} $ & 
$0.34 $ & 
$0.58 $ & 
$0.40 $ \\ 
\hspace{3mm} + Segmentation Map &
$\textbf{1.00} $ & 
$0.44 $ & 
$0.68 $ & 
$0.40 $ \\ 
\hspace{3mm} + Chain-of-Thought  &
$\textbf{1.00} $ & 
$0.36 $ & 
$0.74 $ & 
$0.57 $ \\ 
\hspace{3mm} + Rotation Grid &
$0.99 $ & 
$0.23 $ & 
$\textbf{0.83} $ & 
$\textbf{0.81} $ \\ 
\hspace{3mm} + Rotation Grid Guided &
$0.99 $ & 
$0.31 $ & 
$0.82 $ & 
$0.75 $ \\ 
\hspace{3mm} + \textit{all above} &
$\textbf{1.00} $ & 
$0.28 $ & 
$0.78 $ & 
$0.78 $ \\ 
\bottomrule
\end{tabular}
\caption{Average classification accuracy under different image rotation angles (0°, 90°, 180°, 270°) for \othree{} and \geminitwofivepro{} and various forms of auxiliary information across a single run on \rotsmall{}. Accuracy is scored on a four-way classification task.
}
\label{tab:additional_results_reasoning}
\end{table}

\begin{table*}[ht]
\centering
\small 
\begin{tabular*}{0.75\textwidth}{@{\extracolsep{\fill}} llcccc}
\toprule
\multicolumn{2}{c}{} & \multicolumn{4}{c}{\textbf{Accuracy on Different Degrees of Rotation}} \\
\cmidrule(lr){3-6}
\multicolumn{2}{c}{\textbf{Model}} & \textbf{0°} & \textbf{90°} & \textbf{180°} & \textbf{270°} \\ 
\midrule
{\textit{\smallqwen{}}} \\
Zero-shot & & 
$\textbf{0.60} {\scriptscriptstyle \pm 0.00}$ & 
$\textbf{0.52} {\scriptscriptstyle \pm 0.00}$ & 
$0.16 {\scriptscriptstyle \pm 0.00}$ & 
$0.22 {\scriptscriptstyle \pm 0.00}$ \\ 
\hspace{3mm} + Chain-of-Thought  & & 
$0.18 {\scriptscriptstyle \pm 0.00}$ &
$0.44 {\scriptscriptstyle \pm 0.00}$ & 
$\textbf{0.42} {\scriptscriptstyle \pm 0.00}$ & 
$\textbf{0.28} {\scriptscriptstyle \pm 0.00}$ \\  
\midrule
{\textit{\gglphi{}}} \\
Zero-shot & & 
$\textbf{0.86} {\scriptscriptstyle \pm 0.00}$ & 
$0.10 {\scriptscriptstyle \pm 0.00}$ & 
$0.26 {\scriptscriptstyle \pm 0.00}$ & 
$0.02 {\scriptscriptstyle \pm 0.00}$ \\ 
\hspace{3mm} + Chain-of-Thought  & & 
$0.34 {\scriptscriptstyle \pm 0.00}$ &
$\textbf{0.20} {\scriptscriptstyle \pm 0.00}$ & 
$\textbf{0.40} {\scriptscriptstyle \pm 0.00}$ & 
$\textbf{0.28} {\scriptscriptstyle \pm 0.00}$ \\  
\midrule
{\textit{\gemma{}}} \\
Zero-shot & & 
$\textbf{0.96} {\scriptscriptstyle \pm 0.02}$ & 
$0.18 {\scriptscriptstyle \pm 0.03}$ & 
$\textbf{0.56} {\scriptscriptstyle \pm 0.02}$ & 
$0.18 {\scriptscriptstyle \pm 0.01}$ \\ 
\hspace{3mm} + Chain-of-Thought  & & 
$0.34 {\scriptscriptstyle \pm 0.00}$ &
$0.20 {\scriptscriptstyle \pm 0.04}$ & 
$0.40 {\scriptscriptstyle \pm 0.01}$ & 
$\textbf{0.28} {\scriptscriptstyle \pm 0.05}$ \\  
\midrule
{\textit{\molmo{}}} \\
Zero-shot & & 
$\textbf{1.00} {\scriptscriptstyle \pm 0.00}$ & 
$0.00 {\scriptscriptstyle \pm 0.00}$ & 
$0.00 {\scriptscriptstyle \pm 0.00}$ & 
$0.00 {\scriptscriptstyle \pm 0.00}$ \\ 
\hspace{3mm} + Chain-of-Thought  & & 
$0.58 {\scriptscriptstyle \pm 0.00}$ &
$\textbf{0.28} {\scriptscriptstyle \pm 0.00}$ & 
$\textbf{0.52} {\scriptscriptstyle \pm 0.00}$ & 
$\textbf{0.32} {\scriptscriptstyle \pm 0.00}$ \\  
\midrule
{\textit{\claude{}}} \\ 
Zero-shot & & 
$0.96 {\scriptscriptstyle \pm 0.00}$ & 
$\textbf{0.38} {\scriptscriptstyle \pm 0.00}$ & 
$\textbf{0.48} {\scriptscriptstyle \pm 0.00}$ & 
$\textbf{0.30} {\scriptscriptstyle \pm 0.00}$ \\ 
\hspace{3mm} + Chain-of-Thought  & & 
$\textbf{0.98} {\scriptscriptstyle \pm 0.00}$ & 
$0.28 {\scriptscriptstyle \pm 0.00}$ & 
$0.42 {\scriptscriptstyle \pm 0.00}$ & 
$0.12 {\scriptscriptstyle \pm 0.00}$ \\ 
\bottomrule
\end{tabular*}
\caption{Average classification accuracy under different image rotation angles (0°, 90°, 180°, 270°) for \smallqwen{}, \gglphi{}, \gemma{}, \molmo{}, \claude{} and various forms of auxiliary information across a single run on \rotsmall{}. Accuracy is scored on a four-way classification task.
}
\label{tab:additional_results_other_models}
\end{table*}

\section{Clockwise Prompting}
\label{sec:cw_prompting}

\begin{table*}[ht]
\centering
\small 
\begin{tabular*}{0.75\textwidth}{@{\extracolsep{\fill}} llcccc}
\toprule
\multicolumn{2}{c}{} & \multicolumn{4}{c}{\textbf{Accuracy on Different Degrees of Rotation}} \\
\cmidrule(lr){3-6}
\multicolumn{2}{c}{\textbf{Model}} & \textbf{0°} & \textbf{90°} & \textbf{180°} & \textbf{270°} \\ 
\midrule
{\textit{\qwen{}}} \\
Zero-shot & & 
$\textbf{0.91} {\scriptscriptstyle \pm 0.02}$ & 
$\textbf{0.58} {\scriptscriptstyle \pm 0.03}$ & 
$0.02 {\scriptscriptstyle \pm 0.02}$ & 
$\textbf{0.17} {\scriptscriptstyle \pm 0.01}$ \\ 
\hspace{3mm} + Chain-of-Thought  & & 
$0.82 {\scriptscriptstyle \pm 0.00}$ &
$0.51 {\scriptscriptstyle \pm 0.04}$ & 
$\textbf{0.13} {\scriptscriptstyle \pm 0.01}$ & 
$0.05 {\scriptscriptstyle \pm 0.05}$ \\  
\midrule
{\textit{\gptfour{}}} \\ 
Zero-shot & & 
$0.87 {\scriptscriptstyle \pm 0.02}$ & 
$\textbf{0.66} {\scriptscriptstyle \pm 0.02}$ & 
$0.83 {\scriptscriptstyle \pm 0.02}$ & 
$0.20 {\scriptscriptstyle \pm 0.00}$ \\ 
\hspace{3mm} + Chain-of-Thought  & & 
$\textbf{0.92} {\scriptscriptstyle \pm 0.02}$ & 
$0.63 {\scriptscriptstyle \pm 0.05}$ & 
$\textbf{0.88} {\scriptscriptstyle \pm 0.04}$ & 
$\textbf{0.21} {\scriptscriptstyle \pm 0.05}$ \\ 
\midrule
{\textit{\othree{}}} \\ 
Zero-shot & & 
$\textbf{1.00} {\scriptscriptstyle \pm 0.00}$ & 
$\textbf{0.38} {\scriptscriptstyle \pm 0.06}$ & 
$0.71 {\scriptscriptstyle \pm 0.03}$ & 
$\textbf{0.50} {\scriptscriptstyle \pm 0.03}$ \\ 
\hspace{3mm} + Chain-of-Thought  & & 
$\textbf{1.00} {\scriptscriptstyle \pm 0.00}$ & 
$0.27 {\scriptscriptstyle \pm 0.02}$ & 
$\textbf{0.72} {\scriptscriptstyle \pm 0.02}$ & 
$0.49 {\scriptscriptstyle \pm 0.03}$ \\ 
\bottomrule
\end{tabular*}
\caption{Average classification accuracy under different \textit{clockwise} rotation angles (0°, 90°, 180°, 270°) for \qwen{}, \gptfour{}, \othree{} across two runs on \rotsmall{} using zero-shot or chain-of-thought prompting.
}
\label{tab:clockwise_evals}
\end{table*}

To verify whether MLLMs show a directional preference towards either clockwise or counter-clockwise, we evaluate \qwen{}, \gptfour{}, and \othree{} on \rotsmall{} using \textit{clockwise} angles. Results from all three models closely align with the results from using counter-clockwise angles (\cref{sec:main_results}), indicating the choice of either direction is fully arbitrary. 

\section{Differentiating Clockwise and Counter-clockwise Rotations using \gptfive{}}
\label{sec:gpt5_cw_ccw}

\begin{table}[t]
\centering
\small
\begin{tabular}{lSS}
\toprule
\textbf{GT} $\backslash$ \textbf{Predicted} &
\textbf{Counter-clockwise} & \textbf{Clockwise} \\
\midrule
\multicolumn{3}{l}{\textit{\gptfive{}}} \\
\cmidrule(lr){1-3}
Counter-clockwise & 41 & 9 \\
Clockwise         & 13 & 37 \\
\bottomrule
\end{tabular}
\caption{Accuracy of \gptfive{} in identifying 90° clockwise (CW) versus counter-clockwise rotation (CCW) on \rotsmall{}. Column indicates ground truth rotation (GT). Results are obtained using a temperature of 0.3.
}
\label{tab:gpt5_cw_cww_res}
\end{table}

In \cref{tab:gpt5_cw_cww_res}, we repeat the binary clockwise versus counter-clockwise classification experiment (\cref{sec:cw_ccw_experiment}) using \gptfive{}. \gptfive{} achieves considerably higher accuracy on this binary classification task compared to \gptfour{} and \qwen{}. Notably, performance on 90° and 270° images are significantly higher in the binary task than in the four-way classification task.

This performance difference reveals an important limitation: \gptfive{} frequently misclassifies between portrait rotations (90° and 270°) and landscape rotations (0° and 180°). Despite strong performance on 0° and 180° rotations, this confusion demonstrates that the four-way classification task cannot be effectively reduced to two separate binary classification tasks (one distinguishing between 0° and 180°, and another between 90° and 270°).

\section{Chain-of-Thought Example}
\label{sec:cot_trace}

\begin{figure}[t]
    \centering
    \includegraphics[width=0.5\linewidth]{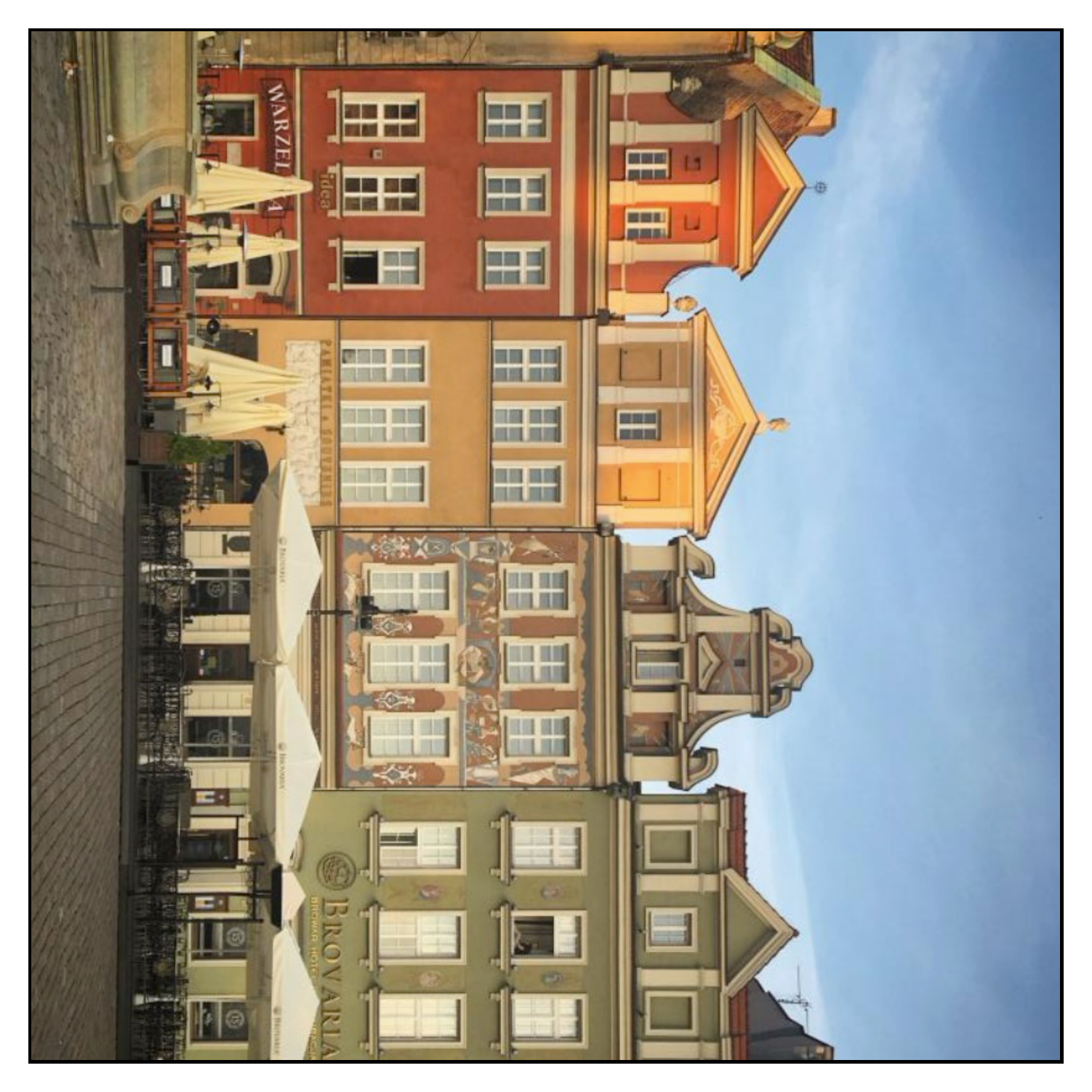}
    \caption{Example image where \gptfour{}'s reasoning falsely distinguishes between two identical forms of rotation.}
    \label{fig:cot_image}
\end{figure}

To further understand how \gptfour{} confuses clockwise and counter-clockwise rotations, we examine the generated reasoning trace of an image in detail. \Cref{fig:cot_image} has been rotated 270° counter-clockwise (or 90° clockwise). However, \gptfour{} generates the following reasoning trace:

\begin{quote}
\ttfamily
\raggedright
To determine the rotation of the image, let's analyze the orientation of the buildings and other elements: [...] 3. **Rotation Analysis**: - A 90-degree rotation counter-clockwise would place the sky on the right and the ground on the left, which matches the current orientation. - A 270-degree rotation clockwise would place the sky on the left and the ground on the right, which does not match. Therefore, the image has been rotated 90 degrees counter-clockwise.
\normalfont
\end{quote}

\noindent As a 90° counter-clockwise rotation is identical to a 270° clockwise rotation, the model falsely distinguishes between the two identical forms of rotation. 

\section{Modifying Temperature}
\label{sec:appx_models}

\begin{table*}[ht]
\centering
\small 
\begin{tabular}{llcccc}
\toprule
\multicolumn{2}{c}{} & \multicolumn{4}{c}{\textbf{Accuracy on Different Degrees of Rotation}} \\
\cmidrule(lr){3-6}
\multicolumn{2}{c}{\textbf{Temperature}} & \textbf{0°} & \textbf{90°} & \textbf{180°} & \textbf{270°} \\ 
\midrule
\multicolumn{6}{l}{\textit{\gptfour{}}} \\
0.0 & & 
$\textbf{0.99} {\scriptscriptstyle \pm 0.00}$ & 
$\textbf{0.69} {\scriptscriptstyle \pm 0.02}$ & 
$0.93 {\scriptscriptstyle \pm 0.00}$ & 
$0.19 {\scriptscriptstyle \pm 0.01}$ \\ 
0.2 & & 
$0.98 {\scriptscriptstyle \pm 0.00}$ & 
$\textbf{0.69} {\scriptscriptstyle \pm 0.02}$ & 
$0.93 {\scriptscriptstyle \pm 0.00}$ & 
$0.2 {\scriptscriptstyle \pm 0.03}$ \\ 
0.5 & & 
$\textbf{0.99} {\scriptscriptstyle \pm 0.00}$ & 
$0.68 {\scriptscriptstyle \pm 0.02}$ & 
$0.93 {\scriptscriptstyle \pm 0.00}$ & 
$0.2 {\scriptscriptstyle \pm 0.03}$ \\ 
0.7 & & 
$0.98 {\scriptscriptstyle \pm 0.01}$ & 
$0.68 {\scriptscriptstyle \pm 0.01}$ & 
$\textbf{0.98} {\scriptscriptstyle \pm 0.00}$ & 
$0.23 {\scriptscriptstyle \pm 0.02}$ \\ 
1.0 & & 
$0.98 {\scriptscriptstyle \pm 0.00}$ & 
$0.65 {\scriptscriptstyle \pm 0.02}$ & 
$0.92 {\scriptscriptstyle \pm 0.00}$ & 
$\textbf{0.25} {\scriptscriptstyle \pm 0.01}$ \\ 
\midrule
\multicolumn{6}{l}{\textit{\qwen{}}} \\
0.0 & & 
$\textbf{0.99} {\scriptscriptstyle \pm 0.00}$ & 
$\textbf{0.51} {\scriptscriptstyle \pm 0.01}$ & 
$0.05 {\scriptscriptstyle \pm 0.01}$ & 
$0.09 {\scriptscriptstyle \pm 0.01}$ \\ 
0.2 & & 
$\textbf{0.99} {\scriptscriptstyle \pm 0.00}$ & 
$0.48 {\scriptscriptstyle \pm 0.00}$ & 
$0.06 {\scriptscriptstyle \pm 0.01}$ & 
$0.11 {\scriptscriptstyle \pm 0.02}$ \\ 
0.5 & & 
$0.94 {\scriptscriptstyle \pm 0.01}$ & 
$0.44 {\scriptscriptstyle \pm 0.00}$ & 
$0.09 {\scriptscriptstyle \pm 0.01}$ & 
$0.19 {\scriptscriptstyle \pm 0.01}$ \\ 
0.7 & & 
$0.9 {\scriptscriptstyle \pm 0.02}$ & 
$0.42 {\scriptscriptstyle \pm 0.02}$ & 
$0.09 {\scriptscriptstyle \pm 0.02}$ & 
$0.2 {\scriptscriptstyle \pm 0.02}$ \\ 
1.0 & & 
$0.83 {\scriptscriptstyle \pm 0.00}$ & 
$0.39 {\scriptscriptstyle \pm 0.00}$ & 
$\textbf{0.13} {\scriptscriptstyle \pm 0.02}$ & 
$\textbf{0.24} {\scriptscriptstyle \pm 0.01}$ \\
\bottomrule
\end{tabular}
\caption{Average classification accuracy under different image rotation angles (0°, 90°, 180°, 270°) for \gptfour{} and \qwen{} using various sampling temperatures. Accuracy is scored on a four-way classification task across three runs on \rotlarge{}.
}
\label{tab:temp_ablation}
\end{table*}

To ensure model performance does not significantly vary with sampling temperature, we examine zero-shot performance of \gptfour{} and \qwen{} (Qwen) on varying temperature settings (\cref{tab:temp_ablation}). We see that performance on all four rotations does not vary significantly for \gptfour{}. Qwen does show slight sensitivity to temperature. As temperature increases, performance on 0° and 90° decreases, while performance on 180° and 270° increases. These results support our primary conclusions, suggesting that the model distinguishes between landscape rotations (0° and 180°) and portrait rotations (90° and 270°). The model defaults to 0° for landscape rotations, and 90° for portrait rotations. With temperature increases, the model's responses begin to resemble random guessing.

\section{\textcolor{black}{The Limitations of Scene Graphs}} 
\label{sec:sg_ablation}

\begin{table}[t]
\centering
\small
\begin{tabular}{c|cccc}
\midrule
 & \textbf{0°} & \textbf{90°} & \textbf{180°} & \textbf{270°} \\
\midrule
\textbf{0°} & - & 0.33 & 0.16 & 0.36 \\
\textbf{90°} & 0.33 & - & 0.21 & 0.20 \\
\textbf{180°} & 0.16 & 0.21 & - & 0.20 \\
\textbf{270°} & 0.36 & 0.20 & 0.20 & - \\
\midrule
\end{tabular}
\caption{Rate of overlapping scene graphs extracted from images rotated different orientations.}
\label{tab:sg_ablation}
\end{table}

Our primary results (\cref{sec:main_results}) show that including various forms of auxiliary information fails to consistently improve model performance. Our intuition behind providing scene graphs is that it would capture unnatural spatial relationships between objects (i.e., “horse above human”). However, strong language priors may override the image understanding, resulting in the model providing the same caption or scene graph regardless of image rotation. To understand why scene graphs fail to improve performance, we examine the overlap between scene graphs extracted from images rotated different orientations. We define two scene graphs as overlapping when both subjects and the predicate are identical, and when both the order of subjects are reversed and the spatial predicate is inverted (e.g., ["cow", "left", "farmer"] and ["farmer", "right", "cow"] are identical scene graphs). 

\cref{tab:sg_ablation} displays the rate of overlapping scene graphs between different image orientations. Noticeably, there is a much higher degree of overlap between 0° and 90°/270° (around 0.3) compared to between 0° and 180° (0.16). These results show the caption model is able to recognize upside-down image and caption accordingly, but fails to correctly capture vertically rotated images.

\section{\textcolor{black}{Alternative Evaluation Methods}} 
\label{sec:alt_setups}

\begin{table}[t]
\centering
\small
\renewcommand{\arraystretch}{1.2}
\begin{tabular}{l r}
\hline
\textbf{Model Name} & \textbf{MAAE ($\downarrow$)} \\
\hline
Qwen 2.5 VL 7B Instruct & $88.72 \pm 2.53$ \\
GPT-4o & $61.09 \pm 0.48$ \\
\hline
\end{tabular}
\caption{Comparison of Mean Absolute Angular Error (MAAE) for \qwen{} and \gptfour{} on a regression-based rotation identification task.}
\label{tab:maae_results}
\end{table}

Our primary experiments (\cref{tab:large_evals}, \cref{tab:small_evals}) both follow the four-option multiple-choice setup widely used in previous literature \citep{anis2025, Joshi2017, udandarao2024visual} to enable a more direct comparison and analysis with previous results. In this section, we explore model performance on alternative evaluation schemes. We see that all alternative setups show inferior performance. We select our current 4-option multi-choice setup for our primary experiments as it shows that, despite the leniency of the setup, frontier models still perform unsatisfactorily.

\myparagraph{Regression.} Given that the current setup already challenges frontier reasoning models, we expect evaluating the models on a regression task on the continuous interval $[0^\circ, 359^\circ]$ will result in further degraded performance. In fact, doing so may actually limit the conclusions and insights we can draw from our experiments as it would be difficult to infer any failure cases without first discretizing the interval. To illustrate this, we provide some further experiments evaluated \qwen{} and \gptfour{} using a modified regression setup. Each image in \rotlarge{} is randomly rotated $0^\circ - 359^\circ$ counter-clockwise. The model is asked to output a single integer representing the degree of rotation. We then calculate the Mean Absolute Angular Error (MAAE) between the estimated and true rotations. Compared to an absolute error, MAAE takes into account the cyclic nature of angular measurements:

\begin{equation*}
\mathrm{MAAE}=\frac{1}{n}\sum_{i}^{n} \min\!\left(|\hat\theta_i-\theta_i|,\,360-|\hat\theta_i-\theta_i|\right)
\label{eq:maae}
\end{equation*}

We find that both models are unable to provide accurate predictions of image orientation. \gptfour{} performed significantly better than \qwen{}, with a difference in MAAE of about 20. Nonetheless, both models averaged $>45$ MAAE across the dataset, indicating a limited ability of current MLLMs to reason over continuous angular spaces.

\begin{table}[t]
\centering
\small
\setlength{\tabcolsep}{8pt} 
\renewcommand{\arraystretch}{1.15}
\begin{tabular}{@{}lc@{}} 
\toprule
\textbf{Number of Bins} & \textbf{Accuracy} \\
\midrule
\textit{Qwen-2.5-VL-7B-Instruct} \\
4  & $0.26 \pm 0.03$ \\
8  & $0.16 \pm 0.01$ \\
10 & $0.15 \pm 0.01$ \\
18 & $0.10 \pm 0.01$ \\
\midrule
\textit{GPT-4o} \\
4  & $0.38 \pm 0.02$ \\
8  & $0.28 \pm 0.01$ \\
10 & $0.23 \pm 0.02$ \\
18 & $0.17 \pm 0.02$ \\
\bottomrule
\end{tabular}
\caption{Classification accuracy on \rotlarge{} using different discretization levels. Rows represent the number of bins, which is equivalent to the number of choices in the multiple-choice question.}
\label{tab:bins_exp}
\end{table}

\myparagraph{Discretization at various granularities.} Alternatively, we evaluate how classification accuracy changes when discretizing the continuous interval $[0^\circ, 359^\circ]$ into various granularities. For each image in \rotlarge{}, we randomly sample a rotation angle in [0°, 359°]. We then discretize the interval into a certain number of bins, assign a letter choice to each bin, and map the ground truth rotation angle to the correct letter choice. The models are then prompted to answer a multiple-choice question, where the number of choices equals the number of bins, and each choice represents a range of angles (e.g., ``A. 0°-20°''). Doing so transformed our previous experiment from a regression task into a classification task. Note that our original experimental setup is identical to the four-bin setup.

We experiment with 4 bins (each representing 90°), 8 bins (45°), 10 bins (36°), and 18 bins (20°). As expected, we see accuracy worsen as we increase the number of bins (\cref{tab:bins_exp}). These results suggest that both evaluated models still lack the ability to reason about fine-grained image orientations.

\section{In-Context Learning} 
\label{sec:icl_experiment}

\begin{table}[t]
\centering
\small
\begin{tabular}{r|rrrr}
\toprule
\textbf{Num. Images} & \textbf{0°} & \textbf{90°} & \textbf{180°} & \textbf{270°} \\
\midrule
0 & 0.99 & 0.51 & 0.05 & \textbf{0.09} \\
8   & \textbf{1.00}  &  0.43 & 0.04 &  0.04 \\
24   & 0.97 &  0.49 & \textbf{0.14} &  0.03 \\
40   & 0.91 &  \textbf{0.59} & 0.09 &  0.06 \\
\bottomrule
\end{tabular}
\caption{\qwen{}'s performance on each orientation when provided different numbers of in-context images.
}
\label{tab:icl_exp}
\end{table}

\myparagraph{Setup.}
To further investigate whether the solution to identifying clockwise and counter-clockwise rotations is simply one of clarifying nomenclature, we implement an in-context learning (ICL) experiment using \qwen{}. Recall \rotlarge{} and \rotsmall{} have an overlap of 25 images. We incrementally sample 2, 6, and 10 distinct in-context examples from the remaining 25 images in \rotsmall{} and evaluate on \rotlarge{}. As each image is rotated in four orientations, the model is shown a total of 8, 24, 40 images. We randomly shuffle the order of in-context images to ensure robustness. Moreover, we provide the ground truth rotation of each in-context example to the model, aiming to guide it towards improved performance.

\myparagraph{Results.}
\Cref{tab:icl_exp} shows the accuracy on each image orientation after injecting in-context examples. We do not see consistent performance improvement, regardless of the number of in-context examples provided. Together with results in \cref{tab:large_evals}, these experiments suggest that various forms of prompt modification are insufficient for identifying rotation. Rather, robust orientation identification demands explicit parameter optimization through fine-tuning.

\section{Similar Images in Spatial-MM}
\label{sec:similar_images_example}

\begin{figure}[ht]
    \centering
    \includegraphics[width=0.4\textwidth]{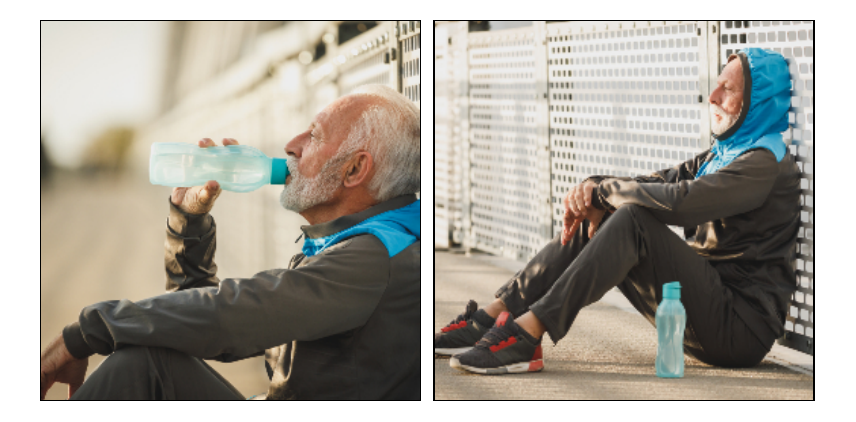}
    \caption{A pair of images in Spatial-MM that closely resemble each other.}
    \label{fig:similar_imgs}
\end{figure}

During the data filtering process, we noted several images in Spatial-MM closely resemble each other (\cref{fig:similar_imgs}). This realization led us to use MS COCO---an out-of-distribution dataset---as the training dataset in our fine-tuning experiment.

\section{\textcolor{black}{Qualitative Examples of Easy and Difficult Images}}
\label{sec:qualitative_examples}

\begin{figure*}[htbp]
    \centering
    \includegraphics[width=0.9\textwidth]{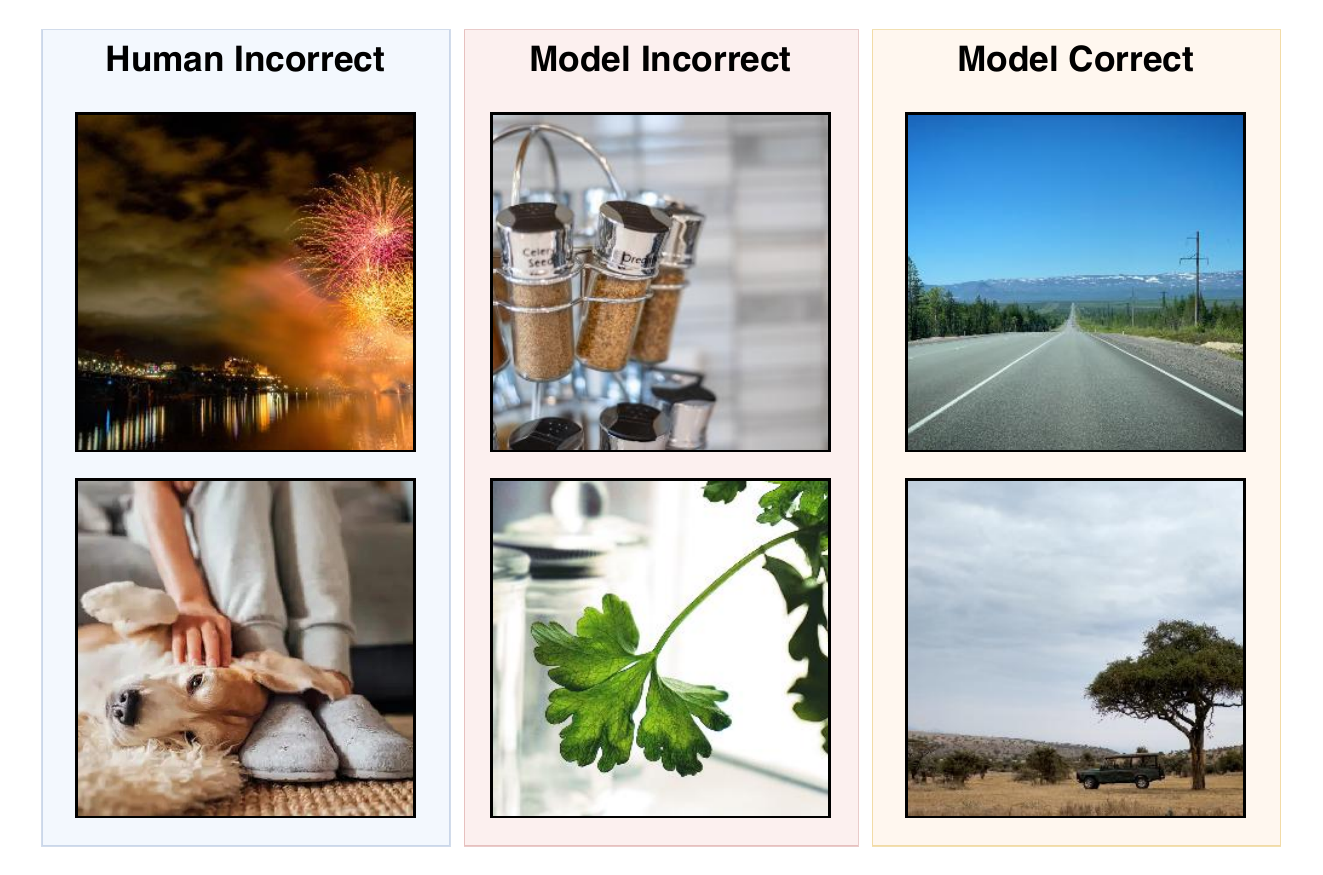}
    \caption{Qualitative examples of images that are difficult for humans (left), difficult for \gptfive{} and \geminitwofivepro{} (center), and easy for both models (right).
    }
    \label{fig:qualitative_ex}
\end{figure*}

This section provides a qualitative illustration of classification difficulty by presenting images that were misclassified by human annotators during Stage 2, alongside examples that both \gptfive{} and \geminitwofivepro{} classified correctly and incorrectly.

\myparagraph{Human errors.} On \rotsmall{}, humans achieve 0.99 accuracy on images rotated 0°, 90°, 180°, and 0.97 accuracy on images rotated 270°. These results indicate that for each rotation, all the annotators combined made fewer than 2 misclassifications. The images that led to misclassifications typically display subjects that can be logically oriented in multiple orientations and require incorporating background information to make an accurate judgment, such as the bottom-left image in \cref{fig:qualitative_ex}.

\myparagraph{Model errors.} To better understand how the models perceive difficulty, we examined the set of images GPT-5 and \geminitwofivepro{} answered correctly when rotated 90°, 180°, and 270° (easy images), and the images they answered incorrectly in all rotations (hard images). We include two example images from each category in the center and right columns of \cref{fig:qualitative_ex}.

We find only 4 images in \rotsmall{} (50 images total) elicit incorrect classifications for both \gptfive{} and \geminitwofivepro{} across the three rotations. The images tend to have these characteristics:

\begin{enumerate}
  \item The image features a prominent primary subject and blurred/subtle background. Moreover, the primary subject can be viable even when rotated in multiple orientations (bottom-center image in \cref{fig:qualitative_ex}). Accurate identification of rotation requires incorporating the background into the reasoning.
  \item The rotational cue is slight or subtle. The ‘Flagged Image’ example in \cref{fig:appendix_sample_imgs} belongs to this category, as the sole rotational cue is the slight tilt of the camera. If the camera was directly looking down, the image would be viable in any rotation.
\end{enumerate}

On the other hand, there are 5 images that both models got correct in all orientations. These images often feature a clear central character, or tend to be natural landscape images featuring clear color distinctions between sky and ground (right column in \cref{fig:qualitative_ex}).

\section{Licenses}
\label{sec:license}

We will publicly release our models, and include our code and data in the supplementary. We provide the following links to the standard licenses for the datasets, code, and models used in this project.

\vspace{0.5em}
\begin{itemize}[leftmargin=*]
    \item \textbf{Spatial-MM:} No license specified (accessed \today). Annotations from \href{https://github.com/FatemehShiri/Spatial-MM}{GitHub.}
    \item \textbf{MS COCO:} \href{https://creativecommons.org/licenses/by/4.0/}{ Creative Commons Attribution 4.0 License.}
    \item \textbf{\qwen{}:} \href{https://github.com/QwenLM/Qwen2.5-VL/blob/main/LICENSE}{Apache~2.0}.
    \item \textbf{\smallqwen{}:} \href{https://huggingface.co/Qwen/Qwen2.5-VL-3B-Instruct/blob/main/LICENSE}{Qwen Research License}.
    \item \textbf{\llama{}:} \href{https://huggingface.co/meta-llama/Llama-3.2-11B-Vision/blob/main/LICENSE.txt}{Llama~3.2 Community License}.
    \item \textbf{\gptfour{}, \gptfourone{}, \gptfive{}, \othree{}:} \href{https://openai.com/policies/services-agreement}{OpenAI Services Agreement} and \href{https://openai.com/policies/service-terms}{Service Terms}.
    \item \textbf{\geminitwoflash{}, \geminitwofiveflash{}, \geminitwofivepro{}:} \href{https://ai.google.dev/gemini-api/terms}{Gemini API Additional Terms of Service}.
    \item \textbf{\gglphi{}:} \href{https://opensource.org/license/mit}{MIT}.
    \item \textbf{\gemma{}:} \href{https://ai.google.dev/gemma/terms}{Gemma Terms of Use}.
    \item \textbf{\claude{}:} \href{https://www.anthropic.com/legal/consumer-terms}{Anthropic Consumer Terms of Service}.
    \item \textbf{\molmo{}:} \href{https://github.com/QwenLM/Qwen2.5-VL/blob/main/LICENSE}{Apache~2.0}.
\end{itemize}

\section{Prompts}
\label{sec:app_prompts}

\cref{fig:subject_prompt} describes the prompt used for extracting primary subjects in images. The list of subjects extracted is later used to obtain bounding boxes, scene graphs, and segmentation maps. \cref{fig:caption_prompt} describes the prompt used for captioning images. \cref{fig:sg_prompt} describes the prompt used for extracting scene graphs from images. \cref{fig:clock_prompt} describes the prompt used for clockwise vs counter-clockwise rotation experiment. \cref{fig:rotation_prompt} describes the system and user prompts for rotation classification. The mapping between letter choice and degrees is shuffled each prompt.

\begin{figure*}[t]
\centering

\begin{UserPrompt}
<Image encoded via. Base64> 

Return a list of objects in this image. The list will later be passed to a bounding box model to extract bounding boxes for each detected object. Format your response as a Python list, surrounded with a Python markdown fence. For example: ```python['fedora', 'woman in green dress', 'man in red suit', ...]``` Each object should have a distinct name. 
ENSURE YOUR RESPONSE FOLLOWS THE FORMATTING REQUIREMENTS!    
\end{UserPrompt}

\caption{Prompts used for extracting primary subjects in images.}
\label{fig:subject_prompt}
\end{figure*}

\begin{figure*}[t]
\centering

\begin{UserPrompt}
<Image encoded via. Base64>

Generate a detailed caption for this image. Do not include any preceding text before the caption. 
\end{UserPrompt}

\caption{Prompts used for captioning images.}
\label{fig:caption_prompt}
\end{figure*}

\begin{figure*}[t]
\centering

\begin{UserPrompt}
<Image encoded via. Base64>

Task: Given the image and key objects, generate a scene graph for this image. Represent each relationship as a three-element tuple with ('subject\_id', 'predicate', 'object\_id'). Extract a set of words describing the location, orientation, directions and spatial or positional relations between key objects in the image. Your answer should be a list of values that are in the format of (object1, relation, object2). The relation MUST be one of [left, right, above, below, facing left, facing right, front, behind]. You are to interpret the image literally. If you see a sky below a mountain, your scene graph must reflect that. Format your response as a Python list of tuples, surrounded by a markdown fence. Example formatting: 

```python
[
("object1", "predicate1", "object2"), 
("object2", "predicate2", "object3"), 
...]
```

Key objects in the image: <previously extracted image subjects>   
\end{UserPrompt}

\caption{Prompts used for extracting scene graphs from images. }
\label{fig:sg_prompt}
\end{figure*}

\begin{figure*}[t]
\centering

\begin{SystemPrompt}
You are an intelligent AI assistant that specializes in identifying rotation in images. You will be given an image that has been rotated 90 or 270 degrees. Specifically, a 90° rotation is a quarter-turn counter-clockwise (the same as 270 degrees clockwise); 270 is three quarter-turns counter-clockwise (the same as 90° clockwise).
\end{SystemPrompt}

\vspace{0.5em}

\begin{UserPrompt}
<Image encoded via. Base64>

Your task is to identify whether the image has been rotated 90 or 270 degrees counter-clockwise. Examine the image closely and identify the rotation. Let's think step-by-step.
\end{UserPrompt}

\caption{Prompts used for clockwise versus counter-clockwise rotation experiment.}
\label{fig:clock_prompt}
\end{figure*}

\begin{figure*}[t]
\centering

\begin{SystemPrompt}
You are an intelligent AI assistant that specializes in identifying rotation in images. You will be given an image and a multiple choice question. Each choice corresponds to the number of degrees the image has been rotated. A 90° rotation is a quarter-turn counter-clockwise; 270° is a quarter-turn clockwise. A 0° rotation indicates the image is right-side up; a 180° rotation indicates the image is upside-down.  
\end{SystemPrompt}

\vspace{0.5em}

\begin{UserPrompt}
<Image encoded via. Base64>

< (if included) Depth map encoded via. Base64>

< (if included) Segmentation map encoded via. Base 64>

Identify whether the image has been rotated.
In addition, you have been provided some extra information about this image below.

< If caption >

The image is given the following caption:
<caption>

< If bounding box> 

Below is the normalized bounding box of objects in the image. Each object is bounded by four floats [xmin, ymin, xmax, ymax] (each float has been normalized between 0 and 1)
<bounding boxes>

< If scene graph > 

Below is a scene graph representing objects within the image and the relationship between them.
<scene graph>

< If depth map >

Attached is also an estimated depth map of the image. The brighter the pixel, the further it is.

< If segmentation map >
Attached is also a segmentation map of the image. Each object has been highlighted a different color.

< If CoT >

What is the rotation of this image? Let's think step-by-step.

< If rotation grid >

Attached is a grid showing the image rotated counter-clockwise in different orientations. The top-left image is the original image shown prior, the top-right image has been rotated 270° counter-clockwise, the bottom-right 180° counter-clockwise, and the bottom-left 270° counter-clockwise. Using these other images as an aid, what is the rotation of the original image? Let's think step-by-step

< If rotation grid guided >

Attached is a grid showing the image rotated counter-clockwise in different orientations. The top-left image is the original image shown prior, the top-right image has been rotated 270° counter-clockwise, the bottom-right 180° counter-clockwise, and the bottom-left 270° counter-clockwise. Use this three step procedure: (1) Carefully examine all four images shown in the grid. (2) Identify the image you are most familiar with, or which image most resembles your training data, as an anchor point. (3) Starting from that image, algebraically determine the rotation of the original image. Using these other images as an aid, what is the rotation of the original image? Let's think step-by-step

< Else > 

Response with a SINGLE LETTER, either A, B, C, or D, representing the correct rotation. You must select one of these choices even if you are uncertain. DO NOT INCLUDE ANYTHING ELSE IN YOUR RESPONSE.

The rotation of the image is:
A. 0
B. 270
C. 90
D. 180

Answer:

\end{UserPrompt}

\caption{System and user prompts for rotation classification.}
\label{fig:rotation_prompt}
\end{figure*}

\end{document}